# SIP-SegNet: A Deep Convolutional Encoder-Decoder Network for Joint Semantic Segmentation and Extraction of Sclera, Iris and Pupil based on Periocular Region Suppression

**Bilal Hassan[1¶*], Ramsha Ahmed[2¶], Taimur Hassan[3], Naoufel Werghi[3]**

## Graphical Abstract

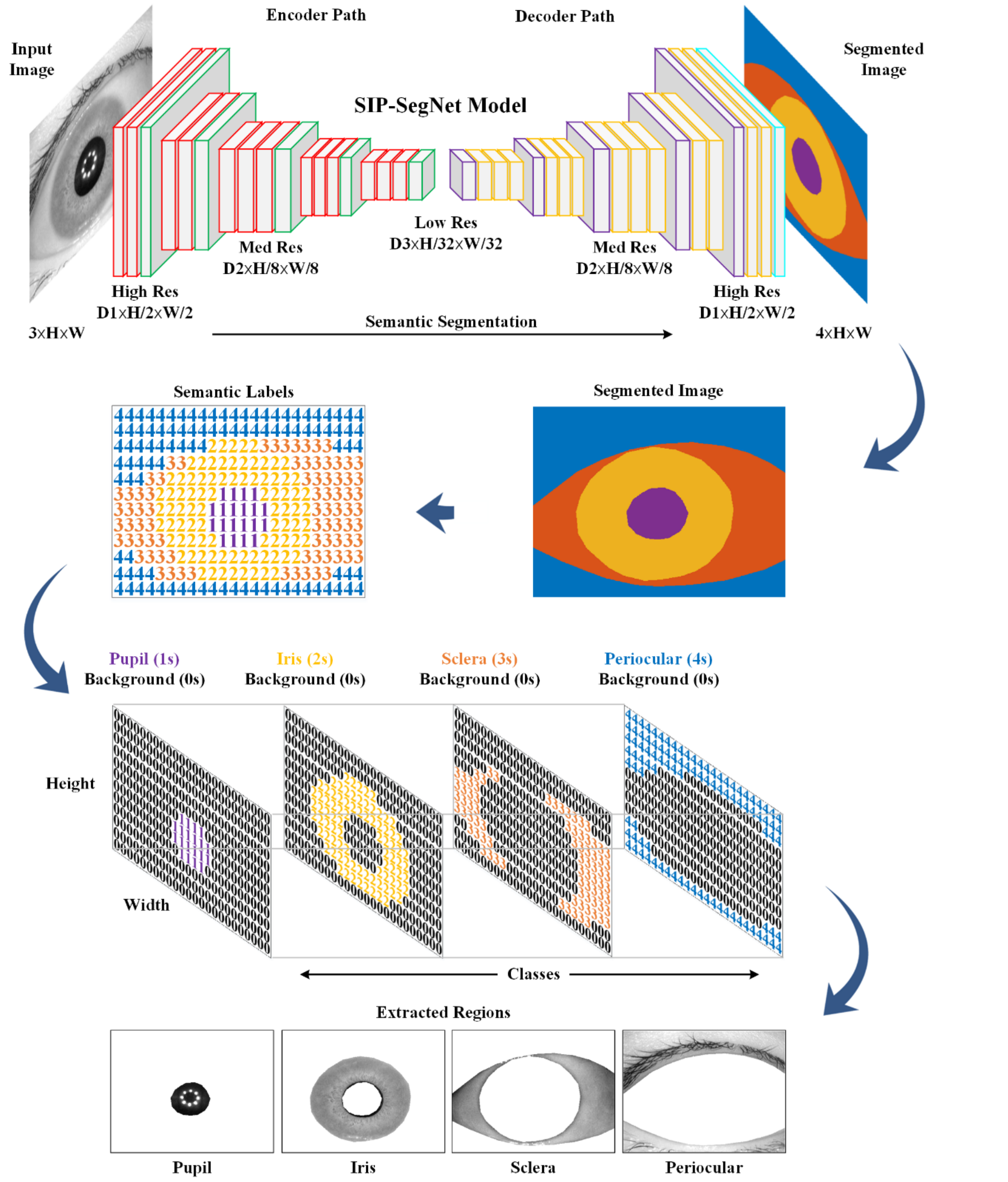

[1]School of Automation Science and Electrical Engineering, Beihang University, Beijing, China
[2]School of Computer and Communication Engineering, University of Science & Technology Beijing, Beijing, China
[3]Center for Cyber-Physical Systems (C2PS), Department of Electrical Engineering and Computer Science, Khalifa University, Abu Dhabi, UAE
¶ These authors contributed equally to this work.

**Abstract:** The current developments in the field of machine vision have opened new vistas towards deploying multimodal biometric recognition systems in various real-world applications. These systems have the ability to deal with the limitations of unimodal biometric systems which are vulnerable to spoofing, noise, non-universality and intra-class variations. In addition, the ocular traits among various biometric traits are preferably used in these recognition systems. Such systems possess high distinctiveness, permanence, and performance while, technologies based on other biometric traits (fingerprints, voice etc.) can be easily compromised. The focus of this work is towards devising a framework to efficiently segment the ocular traits simultaneously, which still remains a challenging task. In this paper, we present a novel deep learning framework called SIP-SegNet, which performs the joint semantic segmentation of ocular traits (sclera, iris and pupil) in unconstrained scenarios with greater accuracy. The acquired images under these scenarios exhibit purkinje reflexes, specular reflections, eye gaze, off-angle shots, low resolution, and various occlusions particularly due to eyelids and eyelashes. These challenging situations limit the performance and reliability of ocular segmentation frameworks. To address these issues, SIP-SegNet begins with denoising the pristine image using denoising convolutional neural network (DnCNN), followed by reflection removal and image enhancement based on contrast limited adaptive histogram equalization (CLAHE). Our proposed framework then extracts the periocular information using adaptive thresholding and employs the fuzzy filtering technique to suppress this information. Finally, the semantic segmentation of sclera, iris and pupil is achieved using the densely connected fully convolutional encoder-decoder network. We used five distinct datasets provided by the Chinese academy of sciences institute of automation (CASIA) to evaluate the performance of SIP-SegNet based on various evaluation metrics. The simulation results validate the optimal segmentation of the proposed SIP-SegNet, with the mean f1 scores of 93.35, 95.11 and 96.69 for the sclera, iris and pupil classes respectively.

*Corresponding author

Email: bilz@live.com

# 1. Introduction

The need for strong authentication systems increased in the last decade due to exponential growth and advancements in the field of information technology. Consequently, researchers are constantly working towards devising secure and reliable authentication systems. In this regard, biometric technology, which refers to authentication of a person based on their measurable physical or behavioral traits, is getting increasingly popular. Moreover, biometric systems are almost impossible to subvert using traditional decryption methods and have unprecedented false match and false rejection rates of 2% [1]. Biometric systems have become an integral part in our daily life because unlike the traditional methods, these systems do not require us to carry or memorize anything such as IDs, pins or passwords [2],[3]. Biometrics are being utilized in a number of critical applications ranging from unlocking the cellphones to withdrawal of cash and from consumer applications to law enforcement and access control in restricted areas [4]-[7].

Person authentication using biometrics are based on various modalities such as: face [8], retina [9], fingerprints [10], facial thermograms [11], vein patterns [12], palm patterns [13] and voice [14]. It has been demonstrated via several studies that ocular traits (sclera, iris, pupil) are preferred over other biometric traits particularly for applications which deem high reliability and accuracy [15]. This is because such systems possess high distinctiveness, permanence, and performance while, technologies based on other traits can easily be compromised such as fingerprints may get burned or affected by allergic skin reactions. Similarly, the performance of the voice recognition system is not reliable as voices can be altered [16].

Sclera, iris and pupil are the three ocular traits. Iris is an annular region between the pupil (black) and the sclera (white) as shown in Fig. 1(a). Eyelids, cornea and aqueous humor naturally protect these traits from the various environmental conditions. Each ocular trait has its own uniqueness and importance as follows:

- **Sclera –** a relatively new biometric trait for person identification has shown promising results [17]-[20]. The vascular pattern in sclera is highly unique for each individual and even observed to be different between the left and right eyes of a person [21] as shown in Fig. 1(a). In addition, it is very hard to counterfeit sclera unlike iris, which can be easily forged by wearing a contact lens [21],[22]. Moreover, segmenting sclera can help in achieving higher accuracy of iris recognitions systems under unconstrained lighting conditions [23].
- **Iris –** the most widely used ocular trait in biometric systems – possesses a high degree of distinctiveness and randomness in terms of its pattern, size, shape and color. This complexity is primarily because of rich and unique textures of iris such as: furrows, rings, freckles, crypts, zigzag or ridges [24] as shown in Fig. 1(a). Moreover, iris trait exhibits a greater immutability throughout a person's life and some studies even concluded the usefulness of iris in postmortem recognition [25]-[27].
- **Pupil –** the least commonly used ocular trait in biometry due to its homogenous structure. There are limited studies based on pupil as a standalone trait for authentication, however, the authors in [28] investigated the size of pupil as a useful biometric trait and their study showed promising results [28]. Usually, segmentation and detection of pupil is considered as the fundamental procedure in developing various computer vision applications such as; iris segmentation [29], eye tracking [30] and gaze estimation [31].

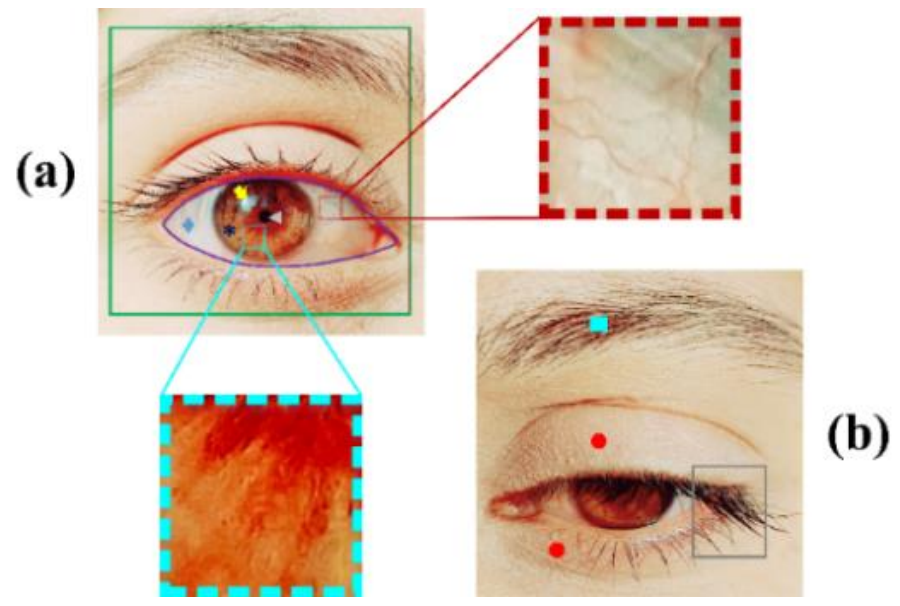

**Fig. 1: Schema of ocular and periocular biometric traits (a) Purple ellipse represents the ocular region with following traits: sclera – marked by (♦), iris – marked by (*), and pupil – marked by (◀). The red and aqua color boxes represent the vascular pattern of sclera and iris texture respectively. Purkinje reflexes are shown by (➡), while the periocular region is indicated in green bounding box (b) Periocular components: eyebrows – marked by (■), eyelids – marked by (●), and eyelashes – marked by gray bounding box.**

The characteristics of these ocular traits in conjunction vary extensively across the human population. Their immutability over time, uniqueness, and randomness can provide a robust and reliable multimodal biometric recognition system. Despite these advantages, joint segmentation of ocular traits is a challenging task mainly due to presence of periocular components as shown in Fig. 1(b). As illustrated from the Fig. 1(b), the performance and reliability of ocular region segmentation framework is limited by the following three major factors:

- **Purkinje reflexes –** resulting from corneal or specular reflections, when the image is acquired under visible-light.
- **Eye gaze –** as a result of body and head movements.
- **Occlusions –** due to eyelids and eyelashes.

According to the above points, the motivation of this work is to perform semantic segmentation of sclera, iris and pupil based on suppression of periocular components. Semantic segmentation is a pixel-wise categorical representation of an image as shown in Fig. 2(c). The performance of semantic segmentation algorithms substantially improved with the advent of deep learning and convolutional neural networks (CNN). In our proposed study, we used the deep learning based semantic segmentation network called Segnet [32].

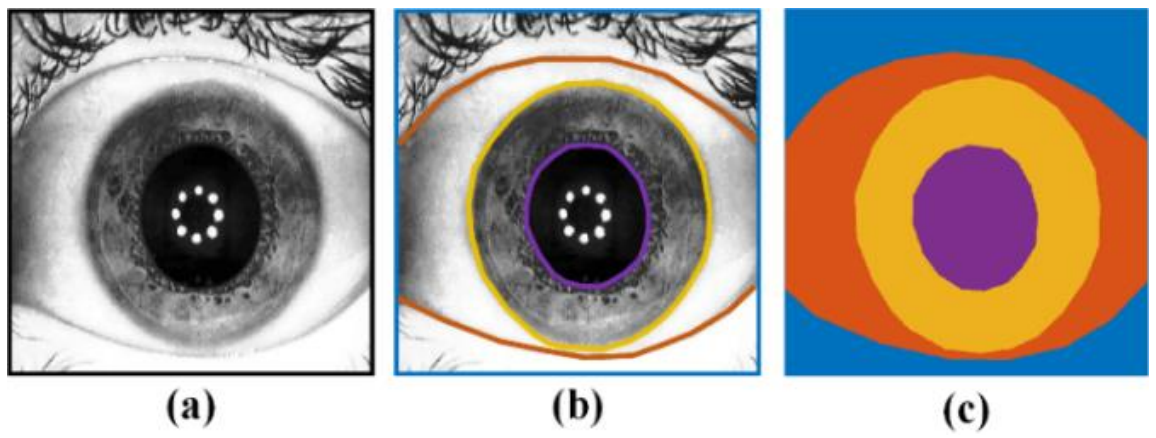

**Fig. 2: (a) Original image (b) Objects localization where blue, orange, yellow and purple color represents the contours of periocular region, sclera, iris and pupil respectively (c) Semantic segmentation.**

## 2. Related Works

### 2.1 State-of-the-art Literature

Most of the work related to ocular biometry in literature is conducted using iris trait. Many researchers proposed different iris recognition methods, most common ones include feature descriptors based methods including integro-differential operator (IDO) [33], Gabor phase-quadrant [34],[35], discrete cosine transforms (DCT) [36], two dimensional discrete wavelet transform (DWT) [37], hierarchical visual codebooks [38], ordinal measures [6], discrete Fourier transforms (DFT) [39] and compressive sensing and sparse coding [40]. In context to iris localization, researchers proposed combination of canny edge detection with circular Hough transform based algorithms [41]. These techniques separately detect each inner and outer boundary of the iris and require an extensive computational time for localization of iris boundaries. Another technique for localization for inner boundary of iris is proposed in [42]. They used a 10×10 vertically sliding rectangular window to locate the pupil first, by finding the black region under the threshold value. The outer iris circle is then found using the difference of sum of greyscale values in a horizontal direction outside the pupil region. Similarly, authors in [43] presented an iris segmentation framework called IrisSeg. Their framework is based on adaptive filtering that adopts a coarse-to-fine strategy to detect the iris boundary.

Deep learning based iris segmentation and recognition methods include the use of fully convolutional networks (FCN) [44]-[46], generative adversarial networks (GAN) [45], fully convolutional encoder–decoder networks (FCEDNs) [47], U-Net and dense U-Net [48]. Authors in [49] proposed an iris segmentation framework at pixel level (PixlSegNet). Their framework is based on the convolutional encoder–decoder architecture, where a stacked hourglass network is used between the encoder and decoder path. Moreover, authors in [50] evaluated the performance of six different pretrained CNN architectures on iris recognition using two publicly available datasets: ND-CrossSensor-2013 and CASIA-Iris-Thousand. They showed that standard CNN features, originally extracted and trained for classification of common objects, can also be transferred and used for recognition of iris. The capsule network based deep learning framework for recognition of iris is proposed in [51]. Their algorithm adjusted the network structure detail to adapt for iris recognition based on modified dynamic routing algorithm within the capsule layers. They employed the transfer learning approach and divided the three pretrained CNN architectures: VGG16, InceptionV3, and ResNet50, into a subnetwork sequences. Primary features are extracted using these series of subnetwork structures as the convolutional part. Three publicly available datasets: JluIrisV3.1, JluIrisV4, and CASIA-V4 Lamp, are used to evaluate the performance of their framework. The proposed framework for iris recognition in [52] is based on multi-layer perceptron neural network–particle swarm optimization (MLPNN–PSO). In their scheme, the iris features are extracted using iris image pre-processing and the two-dimensional (2D) Gabor kernel. The MLPNN architecture comprised of 280 inputs, 1 output neuron along with 100 hidden layer neurons. They used the CASIA-iris V3 and three university of California, Irvine (UCI) datasets for validation purpose and achieved the accuracy rate of 95.36%. IrisConvNet: the deep multimodal-biometric system based on iris recognition is proposed in [15]. They first localized the iris regions in both left and right eyes of the same person, which are then passed to the CNN for extraction of discriminative features and classification using rank fusion technique. They used three publicly available iris datasets: SDUMLA-HMT, CASIA-IrisV3 Interval and IITD, for testing purpose and achieved the accuracy rate of 99.82 for CASIA-IrisV3 and 99.87 for IITD dataset. Fully convolutional deep neural network (FCDNN) framework to segment iris using low quality images is proposed in [7]. They merged four different CNN using semi parallel deep neural network (SPDNN) techniques and used four publicly available datasets: CASIA Thousand, Bath800, UBIRIS v2 and MobBio, for evaluation purpose. Moreover, they also proposed data augmentation technique to produce varied low-quality iris imageries. The reported method for iris localization in [53] is based on AdaBoost algorithm and neural networks, while, the framework in [54] used a combination of feature saliency algorithm and artificial neural network (ANN). The algorithm in [53] classified iris image pixels without any assumption of circularity. They used the university of Bath iris dataset for training and testing purposes, and achieved the accuracy of 98.2%. Authors in [55],[56] introduced the joint fast and cooperative modular neural nets (MNN) for detection of iris in cluttered scenes using a 20×20 pixels window.

Another technique to discriminate the fake iris images towards securing iris recognition systems using multi-patch convolution neural network (MCNN) is proposed in [57]. Their framework learnt mapping function directly between the iris patch pixels and the labels. The final decision is determined by a decision layer based on output of each patch. Authors in [58] recognized the rectangular patterns of iris using self-organizing map neural network. They validated the performance of network on 150 samples of iris and achieved the accuracy of 83%. A deep CNN framework called ContlensNet for detection of iris region with contact lens is proposed in [59]. ContlensNet is a fifteen layers CNN architecture to classify between three different classes: iris with no contact lens, iris with textured contact lens and iris with transparent contact lens. They conducted the experiments on two publicly available datasets: IIIT-Delhi contact lens iris database (IIITD) and Notre Dame cosmetic contact lens database 2013 (ND). Authors in [16] presented a framework for iris recognition based on iris preprocessing and feedforward ANN while the author in [60] used the combination of DCT and ANN for iris classification.

The authors in [21] proposed the convolutional neural network sclera recognition engine (CNNSRE) consisting of four convolutional units and one fully connected unit. They evaluated the framework on sclera segmentation and recognition benchmarking competition (SSRBC) 2015 dataset and received the accuracy of 87.65%. Authors in [61] segmented both iris and sclera using different neural network architectures. They evaluated their algorithm on two different datasets, local and UBIRIS. The method for semantic segmentation of sclera in [23] used the combination of CNN and conditional random fields (CRFs) as a post-processing technique. They used the sclera segmentation and eye recognition benchmarking competition (SSERBC) 2017 dataset for validation purpose of their experiment and received the accuracy of 83.2% in correct classification of sclera pixels. Authors in [62] presented a CNN model called ScleraNET for identification and recognition of person using sclera vasculature pattern. Other deep learning based sclera segmentation frameworks include use of FCN [63], GAN [63] and residual encoder and decoder network [64]. Furthermore, deep learning scheme for recognition of periocular region is proposed in [65]. They separated the ocular and periocular regions first and later by interchanging these regions in different subjects, they created a large set of multi-class artificial samples. These artificial samples are then used for data augmentation and training purpose of CNN.

For segmentation of pupil, authors in [66] used a U-Net based CNN architecture called DeepVOG. They trained the network on two local datasets consisting of 3946 video-oculography (VOG) images. They validated the framework on different datasets and achieved the highest median value of Dice coefficient as 0.978. A multistage CNN architecture for real time segmentation of pupil is proposed in [67],[68]. The first CNN is used for evaluation of downscaled sub regions to estimate rough pupil position. The second stage CNN is used to refine the pupil position using initial estimates and its surrounding sub regions. Semantic segmentation and detection of pupil using deconvolutional neural network is proposed in [69]. They used 24 different datasets consisting of overall 94,161 images for training and testing purposes of their algorithm. Authors in [70] estimated the pupil center based on AlexNet architecture using image patch classification, while authors in [71] used U-Net for the pupil segmentation. The approach in [71] is validated on two different datasets, UBIRIS consisting of close-up view of eye and Gi4E containing full facial image. In case of Gi4E dataset, they first extracted the eye region using dlib [72], which is then passed to the pupil segmentation network. Another scheme based on the combination of pulling and pushing method and a batch self-organizing maps (BSOM) neural network for segmentation of pupil is proposed in [73],[74]. They initially estimated the pupil contour using modified pulling and pushing method and then the exact pupil shape is adjusted using the BSOM network. Two end to end CNN architectures inspired from faster region based convolutional neural network (RCNN) and single shot multibox detector (SSD) for joint segmentation of iris and pupil are proposed in [75]. These networks produced two separate circular region of interests corresponding to iris and pupil from an input eye image. Another deep convolutional neural network called DeepEye for detection of pupil based on atrous convolutions and spatial pyramids is proposed in [76].

## 2.2 Contributions

Our research focuses on the semantic segmentation of three ocular traits: sclera, iris and pupil using a deep convolutional encoder-decoder network known as Segnet [32]. The notable contributions of this research are summarized as under:

1. An efficient deep learning based semantic segmentation framework called SIP-SegNet is proposed. SIP-SegNet is a unique and novel framework that can perform joint segmentation of multiple ocular traits: sclera, iris and pupil. To the best of authors knowledge, this is the first work that investigates the potential use of deep learning based semantic segmentation approach towards the segmentation of biometric traits. Moreover, our proposed framework is the first of its kind to perform the concurrent segmentation of sclera, iris and pupil. In the past, researchers have mainly focused on segmenting a single ocular trait or two but no unique algorithm is reported which can jointly segment these ocular traits.
2. In our proposed framework for segmentation of ocular traits, one of the fundamental task is to suppress the periocular region information such as eyelashes, eyebrows and spectacles etc. In this context, we developed a technique based on optimal adaptive thresholding to extract the components in the periocular region. The similar technique can be extended towards devising an algorithm for covert recognition of humans, which is currently an active research area in the field of security and forensics applications [77].
3. Our framework incorporates an efficient image denoising scheme based on residual learning [78]. This is the first time DnCNN is used to denoise iris images. Our findings indicated that the residual learning property of DnCNN can be extremely useful for removing various artifacts in iris imagery such as: specular reflection, ocular motion (gaze) etc. In addition, it can also be utilized to correct the poorly focused images.
4. We tested the performance of our proposed framework on five publicly available datasets of CASIA-IrisV4 database containing a total of 52,034 images, which are collected under different environmental conditions. The performance of our framework is demonstrated using various evaluation metrics. The simulation results presented validate the excellent performance of our proposed framework in ocular segmentation.
5. To conclude, the proposed framework is devised to extract the highly unique ocular traits, which can be used in conjunction with existing state-of-the-art algorithms towards developing a robust multimodal biometric recognition system.

The remaining paper is organized as follows: section 3 is about the description of datasets used in this research, section 4 indicates the annotation strategy utilized to label the ground truths. Section 5 explains the proposed framework in detail whereas, in section 6 we introduced the experimental setup and various evaluation metrics. Results are presented in section 7 followed by discussion in section 8. Finally, section 9 concludes the paper. Table I provides a list of major acronyms along with their definitions used in this paper in order to assist the readers. Moreover, Table II defines various symbols and notations that are used in our study.

**Table I: List of Acronyms.**

| Acronym | Definition | Acronym | Definition |
|---|---|---|---|
| ATMED | Asymmetrical Triangular Fuzzy Filter with Median Center | FNR | False Negative Rate |
| AUC | Area Under Curve | FPR | False Positive Rate |
| BN | Batch Normalization | FWIoU | Frequency Weighted Intersection over Union |
| BRISQUE | Blind/ Referenceless Image Spatial Quality Evaluator | MIoU | Mean Intersection over Union |
| CASIA | Chinese Academy of Sciences Institute of Automation | NIQE | Naturalness Image Quality Evaluator |
| CII | CASIA Iris Interval | NLM | Non Local Means |
| CIL | CASIA Iris Lamp | NPV | Negative Predictive Value |
| CIS | CASIA Iris Syn | PIQE | Perception based Image Quality Evaluator |
| CIT | CASIA Iris Thousand | PSNR | Peak Signal to Noise Ratio |
| CITW | CASIA Iris Twin | ReLU | Rectified Linear Unit |
| CLAHE | Contrast Limited Adaptive Histogram Equalization | ROC | Receiver Operating Characteristics |
| DIO | Daughman’s Integro-differential Operator | SGDM | Stochastic Gradient Descent with Momentum |
| DnCNN | Denoising Convolutional Neural Network | SSIM | Structural Similarity Index |

**Table II: Symbols and Notations.**

| Symbols | Definition | Symbols | Definition |
|---|---|---|---|
| $\mathcal{Y}$ | Degraded Image | $\mathfrak{F}$ | Fuzzified Image using ATMED Fuzzy Filtering |
| $\mathcal{X}$ | Pristine Image | $\mathfrak{P}$ | Preprocessed Image |
| $\mathcal{N}$ | Noise | $A$ | Accuracy |
| $\mathcal{R}$ | Residual Image | $P$ | Precision |
| $\mathcal{X}'$ | Restored Image | $R$ | Recall |
| $\mathcal{L}$ | Network Loss Function | $S$ | Specificity |
| $\mathfrak{I}$ | Intensity Transformed Image | $MA$ | Mean Accuracy |
| $\mathcal{D}$ | Denoised Image using DnCNN | $GA$ | Global Accuracy |
| $\mathcal{E}$ | Enhanced Image using CLAHE | $D$ | Dice Similarity Coefficient |
| $\mathcal{F}$ | Filtered Image after Reflection Removal | $F1$ | F1 Boundary Score |
| $\mathcal{P}'$ | Binary Enhanced Image using Adaptive Thresholding | $N1$ | Nice1 |
| $\mathcal{P}$ | Periocular Image after Subtracting Pupil | $N2$ | Nice2 |

# 3. Dataset Details

## 3.1 Existing Ocular Datasets

There are various ocular datasets that are publicly available to conduct research in the area of ocular biometrics [79], where the focus of majority of these datasets is towards the most dominant ocular trait – the iris [80]-[91]. These datasets are also well suited for segmentation, recognition or detection of pupil because of its homogenous black structure, however, these listed datasets pose a major limitation for research into sclera recognition. This is because these datasets are commonly acquired in the near-infrared (NIR) spectrum, which depletes the discernible discriminative sclera vasculature pattern.

Table III summarizes the characteristics of widely used ocular datasets. We have opted for CASIA database in our study mainly because: (i) it is one of the largest datasets with a lot more subjects and intra-class variations and (ii) our prime focus is towards the segmentation and extraction of ocular modalities for which this database is well suited.

**Table III: Popular Databases used in Ocular Biometrics.**

| Database | Modality | Spectrum | Subjects | Images | Public |
|---|---|---|---|---|---|
| CASIA v4 [80] | I | NIR | > 2800 | 54601 | Yes |
| UBIRIS v1 [81] | S, I | VIS | 241 | 1877 | Yes |
| UBIRIS v2 [82] | S, I, PO | VIS | 261 | 11102 | Yes |
| IITD [83] | I | NIR | 224 | 1120 | Yes |
| SBVPI [84] | S, I, PO | VIS | 55 | 1858 | Yes |
| ND-IRIS-0405 [85] | I | NIR | 356 | 64980 | Yes |
| IMP [86] | PO | Both | 62 | 930 | Yes |
| UTIRIS [87] | S, I | Both | 79 | 1540 | Yes |
| UBIPr [88] | PO | VIS | 261 | 10950 | Yes |
| IUPUI [89] | S, I, PO | Both | 44 | 352 | No |
| MICHE-I [90] | S, I, PO | VIS | 92 | 3732 | Yes |
| MASD [91] | S | VIS | 82 | 2624 | No |

S= Sclera, I= Iris, PO= Periocular, NIR= Near-Infrared, VIS= Visible, CASIA v4 [80] is the iris image database provided by the Chinese Academy of Sciences Institute of Automation; UBIRIS v1 [81] and UBIRIS v2 [82] are the ocular while UBIPr [88] is the periocular region databases provided by the University of Beira Interior; IITD [83] is the dataset provided by the Indian Institute of Technology Delhi; SBVPI – Sclera Blood Vessels, Periocular and Iris is a multipurpose dataset provided by the University of Ljubljana [84]; ND-IRIS-0405 [85] is a dataset consisting of iris images collected by the University of Notre Dame; IMP [86] is the Multispectral Periocular dataset provided by the Indraprastha Institute of Information Technology, Delhi (IIITD); UTIRIS [87] is the iris image repository provided by the University of Tehran; IUPUI [89] is the multi-wavelength database provided by the Indiana University - Purdue University Indianapolis; MICHE-I – Mobile Iris Challenge Evaluation is the iris bio-metric dataset provided by the University of Salerno [90]; MASD – Multi-Angle Sclera Dataset is used in Sclera Segmentation Benchmarking Competition (SSBC) [91].

## 3.2 Dataset used in Proposed Study

In this study, the datasets used for training, validation and testing purposes are collected and arranged by the Chinese academy of sciences institute of automation [80]. We used the five subsets of CASIA-IrisV4 database: (i) CASIA-iris-interval (CII), (ii) CASIA-iris-syn (CIS), (iii) CASIA-iris-lamp (CIL), (iv) CASIA-iris-thousand (CIT) and (v) CASIA-iris-twins (CITW). The five subsets contain a total of 52,034 iris images of about 2800 subjects. All images are captured under the near-infrared or synthesized illumination and are in the JPEG format with a grayscale depth of 8 bits. Table IV shows the statistics and features of each of the five subsets of CASIA-IrisV4 database used in this study. The sample images of each subset are shown in Fig. 3.

**Table IV: Dataset Details.**

| Subsets | Images | Subjects | Resolution | Environment | Sensor | Features |
|---|---|---|---|---|---|---|
| **CII** | 2639 | 249 | 320×280 | Indoor | CASIA NIR LED | Cross-session extremely clear iris images<br>Detailed iris texture features |
| **CIS** | 10000 | 1000 | 640×480 | - | - | Synthesized iris images |
| **CIL** | 16212 | 411 | 640×480 | Indoor | OKI IRISPASS-h | More intra-class variations<br>Elastic deformation of iris texture<br>Non-linear iris normalization |
| **CITW** | 3183 | 200 | 640×480 | Outdoor | OKI IRISPASS-h | Iris images of 100 pairs of twins |
| **CIT** | 20000 | 1000 | 640×480 | Indoor | Irisking IKEMB-100 | High-quality iris image<br>Intra-class variations due to eyeglasses and specular reflections. |

CII= CASIA-Iris-Interval, CIS= CASIA-Iris-Syn, CIL= CASIA-Iris-Lamp, CITW= CASIA-Iris-Twins, CIT= CASIA-Iris-Thousand.

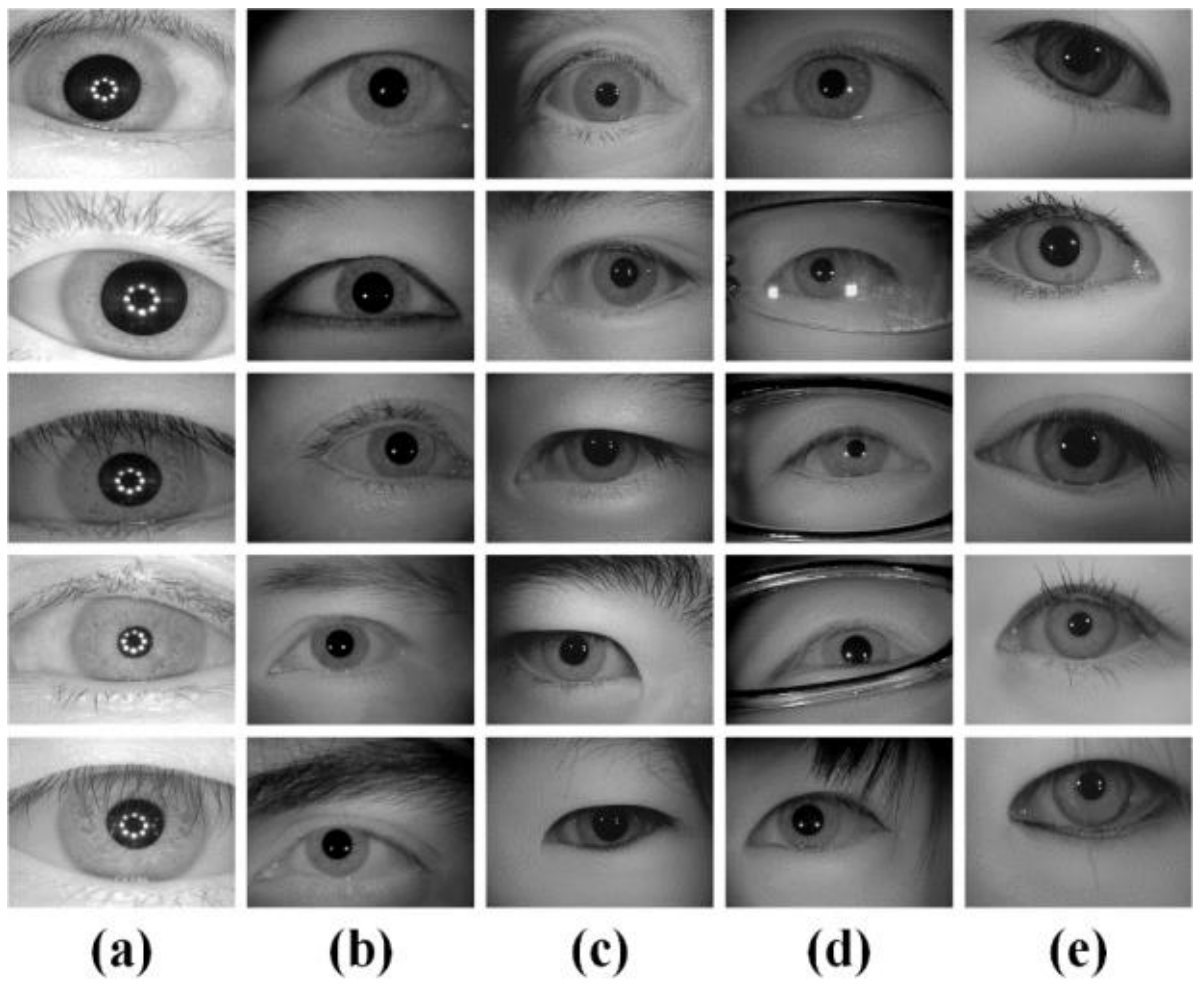

**Fig. 3: Sample images (a) CASIA-iris-interval (b) CASIA-iris-syn (c) CASIA-iris-lamp (d) CASIA-iris-thousand (e) CASIA-iris-twins.**

# 4. Annotation Method

Good ground truth data is vital for evaluating the performance of automated algorithms. Training the proposed model requires the pixel-level ground truths of periocular (background), sclera, iris and pupil region. Unfortunately, CASIA database does not contain the ground truths. Therefore, we used the interactive image labeler application provided by mathworks for labeling the entire dataset and exporting the label data for training. It allows to annotate objects as pixels for semantic segmentation, rectangular region of interest (ROI) for object detection, and scenes for image classification.

Manually labeling the large datasets from scratch require a substantial amount of time and resources, therefore, we automated the labeling of ground truth data by importing the pretrained semantic segmentation framework and fine-tuned its parameters into the workflow of the image labeler application. We then refined the ground truth data manually to correct the false labels against some of the challenging images. Fig. 4 shows the manually labeled ground truths for each of the five subsets of CASIA-IrisV4 database, where the ground truth labels for the periocular region, sclera, iris and pupil are indicated in blue, orange, yellow and purple color respectively. In our study, tear duct is considered as a part of sclera ground truth label.

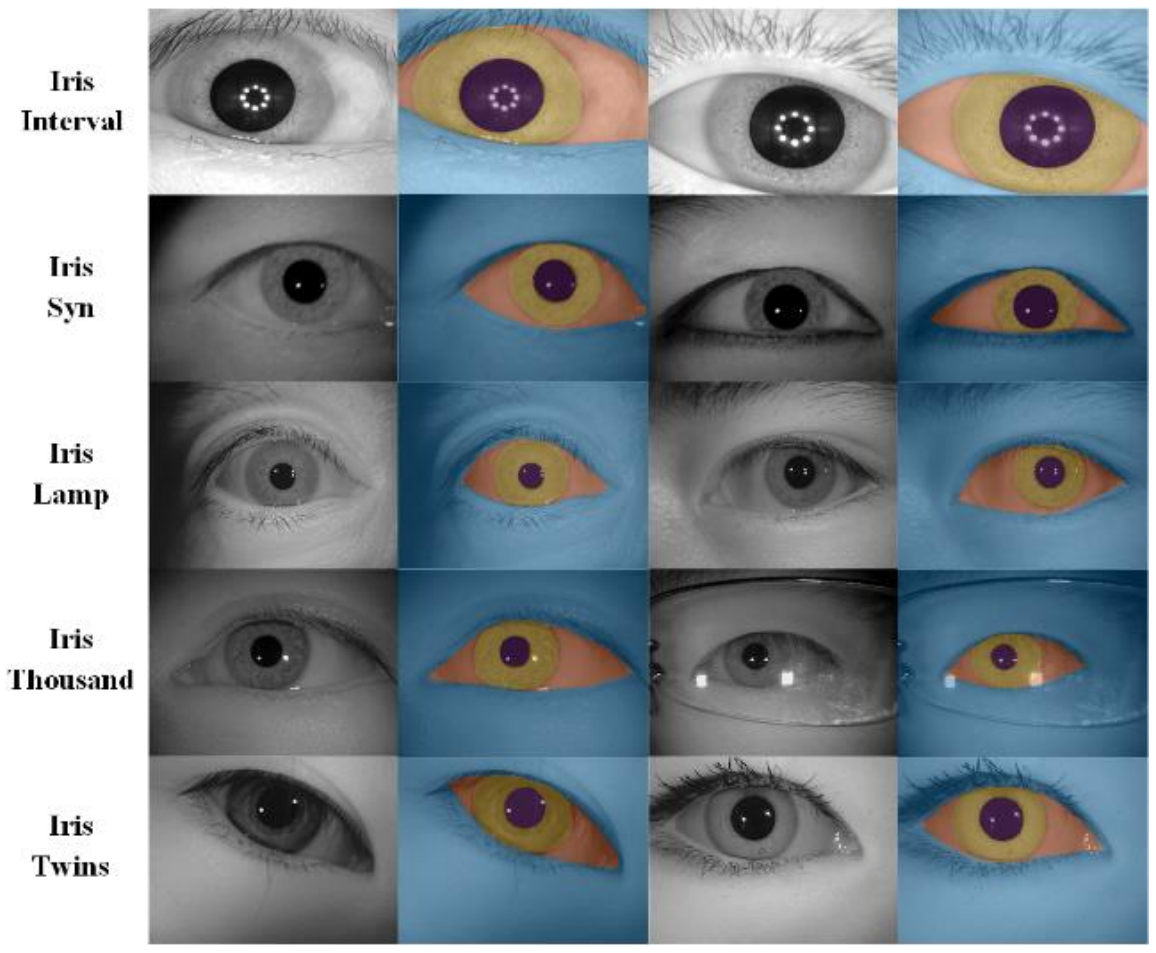

**Fig. 4: Sample images and their respective ground truth labels for each dataset.**

## 5. Proposed Framework

SIP-SegNet framework consists of six main stages as shown in Fig. 5. The preprocessing stage in our proposed framework denoises, enhances and suppresses the periocular information in the pristine images. In the next stage, the ground truths are generated using the preprocessed images. Each image consists of 4 ground truth labels: (i) periocular is represented in blue, (ii) sclera is represented in orange, (iii) iris is represented in yellow and (iv) pupil is represented in purple color. Using both the preprocessed images and their corresponding ground truth labels, we split the data into training, validation and testing sets during the data preparation stage. In the data augmentation stage, we performed various augmentations on the training set to create the augmented datastore for network training. It is followed by the class balancing stage, where we assigned the weights to under-represented classes for optimal network training. In the final stage, we trained SIP-SegNet to perform semantic segmentation and tested its performance using various evaluation metrics as discussed in the following sections.

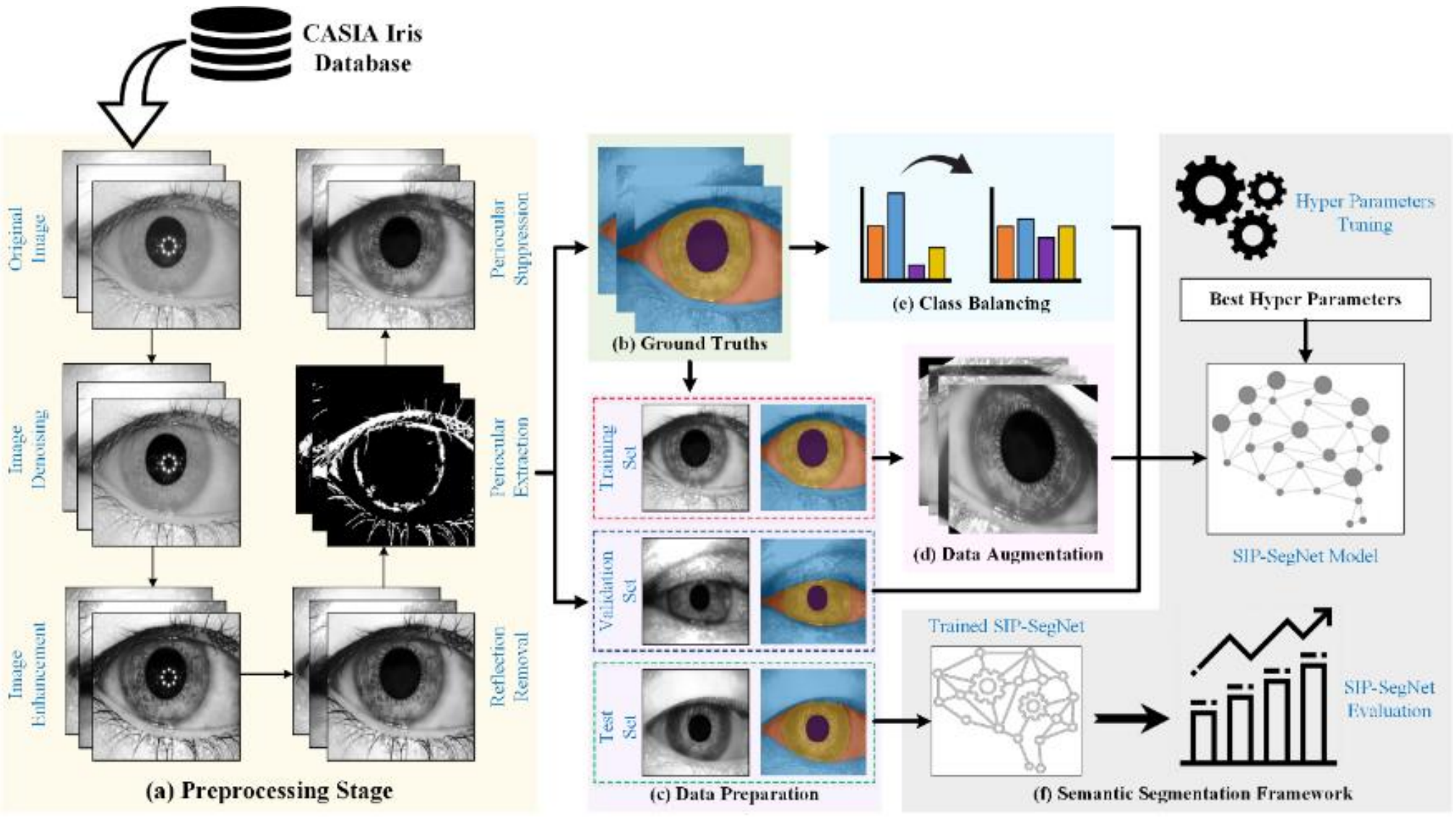

**Fig. 5: Schema of proposed SIP-SegNet semantic segmentation framework.**

## 5.1 Preprocessing Stage

### 5.1.1 Image Denoising

In our study, we are using five different subsets of CASIA-IrisV4 database [80], containing a total of 52046 eye scans. The attributes of these five subsets vary extensively from one another in terms of environment (indoor/outdoor), illumination, camera sensor and resolution [80] as shown in Table IV. Hence, before feeding to the CNN architecture, image handling and processing is required to scale and convert these different type of images to a standard image according to the dimension, activation shape and size specified by the CNN input layer. These image handling operations may result in the loss of information and different artifacts may also appear in the processed image, when compared with the pristine image. Therefore, in the preprocessing stage of our proposed system, we first denoised the pristine images using feed-forward DnCNN. Afterwards, we applied the intensity transformation and contrast stretching techniques to further enhance the denoised scan as explained in the subsections.

#### 5.1.1.1 DnCNN

DnCNN is a discriminative model which utilizes the residual learning and batch normalization to restore the degraded scan [78]. In addition to general denoising tasks such as Gaussian noise removal, the residual learning property of DnCNN also allows it to deal with the more complicated tasks such as compressed images deblocking and single image super resolution (SISR). According to the general image degradation model, $\mathcal{Y}$ is the degraded scan resulting from the addition of the noise $\mathcal{N}$ in the pristine image $\mathcal{X}$ as expressed in Eq. (1):

$$\mathcal{Y}(i,j) = \mathcal{X}(i,j) + \mathcal{N}(i,j) \tag{1}$$

This noisy image $\mathcal{Y}$ is input to the DnCNN network, trained to estimate the residual image $\mathcal{R}$, resulting from the difference between the original image $\mathcal{X}$ and the degraded image $\mathcal{Y}$ as expressed in Eq. (2):

$$\mathcal{R}(\mathcal{Y}) = \mathcal{Y}(i,j) - \mathcal{X}(i,j) \tag{2}$$

Since, DnCNN is based on residual learning formulation, it considers noise $\mathcal{N}$ in Eq. (1) to be approximately equal to the residual image $\mathcal{R}$ containing information about the pristine image degradation as expressed in the Eq. (3):

$$\mathcal{R}(\mathcal{Y}) \approx \mathcal{N}(i,j) \tag{3}$$

Once the network is trained, the undistorted version of the degraded image $\mathcal{X}'$ can be reconstructed by subtracting the residual image from the degraded image as expressed in Eq. (4):

$$\mathcal{X}'(i,j) = \mathcal{Y}(i,j) - \mathcal{R}(\mathcal{Y}) \tag{4}$$

##### 5.1.1.1.1 Training Data

For training the DnCNN, we used a batch of 1500 images by randomly selecting 300 images from each of the five subsets of CASIA-IrisV4 database. Firstly, we generated two separate datastores: (i) degraded image datastore, which is created by adding the random combination of the four image distorting factors; Gaussian noise, scaling, rotation, and motion blur to the original image $\mathcal{X}$. (ii) residual image datastore, which is created by subtracting the original image $\mathcal{X}$ from the degraded image $\mathcal{Y}$ as expressed in Eq. (2). Table V lists the description of datastores used to train the denosing network. The details of image distortion parameters with their value ranges for generation of degraded images are shown in Table VI. Fig. 6 shows the sample original, degraded and residual images from each of the five subsets of CASIA-IrisV4 database.

**Table V: Description of Training Datastores.**

| CASIA Subsets | No. of Scans in Each Datastore |
|---|---|
| Iris-Interval | 300 |
| Iris-Syn | 300 |
| Iris-Lamp | 300 |
| Iris-Thousand | 300 |
| Iris-Twins | 300 |
| **Total** | **1500** |

**Table VI: Image Distortion Parameters.**

| Distortion Parameter | Value (Range) |
|---|---|
| Gaussian Noise | 0 Mean,<br>(0.005 to 0.015) Variance |
| Scaling | (1.05 to 1.1) Percent |
| Rotation | (-05 to 05) Degree |
| Motion Blur | Horizontal (1 to 9) Pixels |
| | Vertical (1 to 9) Pixels |
| | Theta (-20 to 20) Degrees |

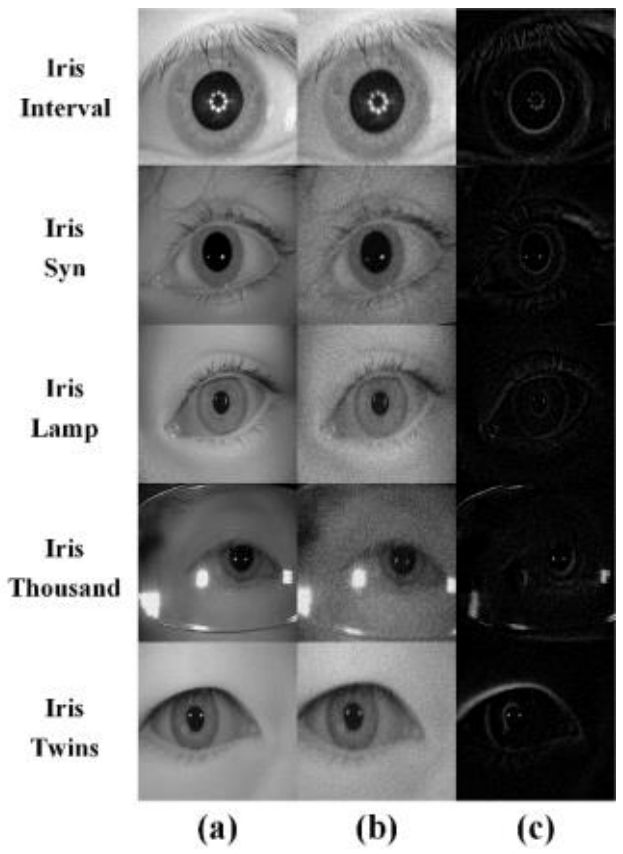

**Fig. 6: (a) Original image (b) Degraded image (c) Residual image.**

### 5.1.1.1.2 Extraction of Patches

After creating the two image datastores, we randomly extracted the corresponding 50×50 sized patches from both the datastores and stored them in a new datastore called patch extraction datastore. The dimensions of the patch are selected in a way that it remains compatible with the input layer specification of DnCNN. We specified the mini-batch size of 64, so each image generated 64 random patches of size 50×50. This new datastore is used for training the network and provides the mini-batches of data at each iteration.

### 5.1.1.1.3 Network Architecture and Training Parameters

DnCNN is the 59 layered regression architecture having a network depth of 20 (convolutional layers) [78]. We trained the network using the patch extraction datastore as explained earlier. DnCNN uses the half mean squared error as a loss function between the residual images $\mathcal{R}$ and estimated ones from the noisy input $\mathcal{T}$ over the total number of training patches $N$ as expressed in Eq. (5):

$$\mathcal{L} = \frac{1}{2N} \sum_{i=1}^{N} (\mathcal{R}_i - \mathcal{T}_i) \quad (5)$$

The network parameters during the training phase have been updated using the stochastic gradient descent with momentum (SGDM) optimizer. Table VII shows the details of both mini-batch and optimizer hyper parameters used to train the DnCNN model in our proposed study. The training performance of DnCNN is shown in Fig. 7.

**Table VII: DnCNN Training Hyper parameters.**

| Parameters | Value |
|---|---|
| Training Images | 1500 |
| Patch Size | 50 x 50 |
| Patches/ Image | 32 |
| Total Patches | 48000 |
| Maximum Epochs | 30 |
| Mini-Batch Size | 64 |
| Iterations/ Epoch | 750 |
| Total Iterations | 22500 |
| Solver | SGDM |
| Momentum | 0.9 |
| Learning Rate | 0.1 |
| Gradient Threshold | 0.005 |
| Weight Decay | 0.0001 |

SGDM= Stochastic Gradient Descent with Momentum.

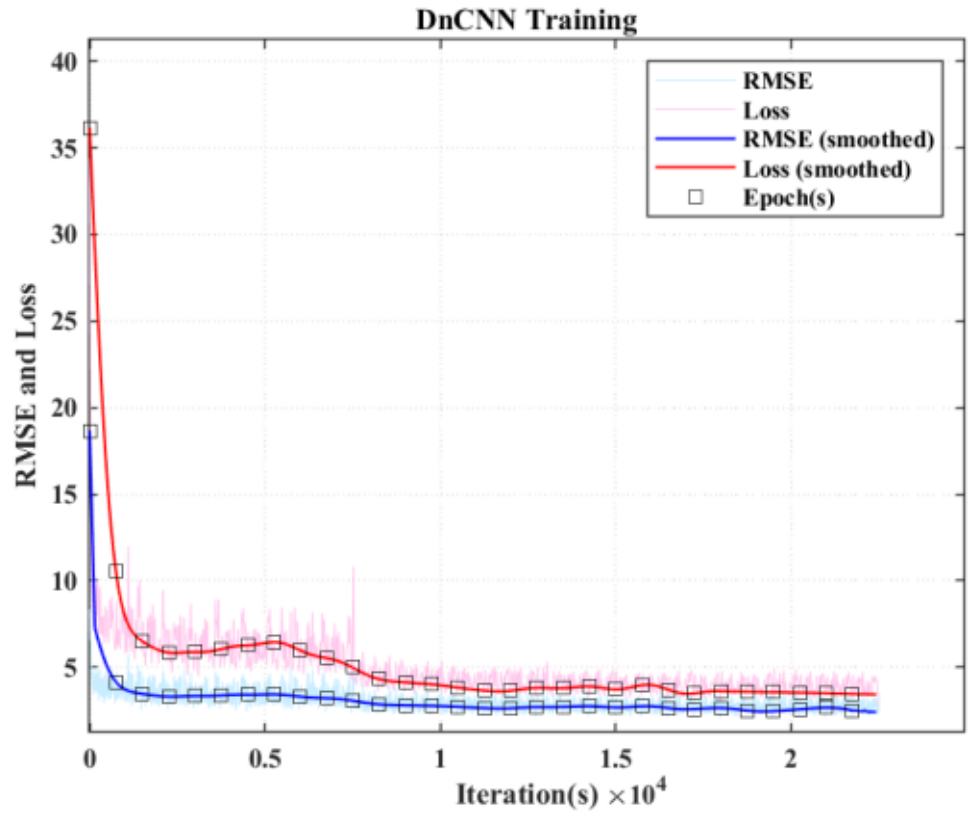

**Fig. 7: DnCNN training performance.**

### 5.1.1.1.4 Objective Evaluation of Denoised Images

The performance of our denoising network is evaluated using set of both reference and non-reference based image quality metrics. Since, the ideal undistorted images are unavailable, we considered the original images as a reference to quantify the following two reference based quality metrics: (i) peak signal-to-noise ratio (PSNR) and (ii) structural similarity index (SSIM) [92].

Additionally, we computed the denoised image quality scores based on expected image statistics using the following three non-reference based quality metrics: (i) naturalness image quality evaluator (NIQE) [93], (ii) blind/ referenceless image spatial quality evaluator (BRISQUE) [94] and (iii) perception based image quality evaluator (PIQE) [95]. These three methods compare the input image to a default model and a smaller value of these metrics indicate a better image quality. Table VIII shows the average values of the evaluation metrics for each of the five subsets of CASIA-IrisV4 database. Also, we have computed the overall mean value of each metric over the complete database as shown in Table VIII.

**Table VIII: Objective Evaluation of Denoised Images.**

| Metric | Dataset: Five Subsets of CASIA-IrisV4 database | | | | | Overall Mean Value |
|---|---|---|---|---|---|---|
| | CII* | CIS* | CIL* | CITW* | CIT* | |
| **PSNR** | 39.80 | 43.11 | 49.74 | 41.59 | 38.52 | **42.55** |
| **SSIM** | 0.90 | 0.92 | 0.95 | 0.91 | 0.89 | **0.91** |
| **NIQE** | 4.16 | 3.85 | 3.21 | 4.02 | 4.27 | **3.90** |
| **BRISQUE** | 24.86 | 22.69 | 20.65 | 22.84 | 25.11 | **23.23** |
| **PIQE** | 26.61 | 25.15 | 22.96 | 25.46 | 27.08 | **25.45** |

*represents the average value of each metric computed against the total images present in each subset.

### 5.1.1.2 Intensity Transformation

Intensity transformation is a useful image enhancement technique especially when dealing with the grayscale images. It is a way to transform each intensity value in an input image to the corresponding output value as expressed in Eq. (6):

$$\Im(i,j) = T[\mathcal{X}(i,j)] \tag{6}$$

Where, $\mathcal{X}$ is the input image, which is the denoised image using DnCNN. $\Im$ is the output image containing new corresponding intensity values at points $(i,j)$ and $T$ is the intensity transformation function.

CASIA-IrisV4 is a grayscale database and intensity values in some of its subsets required to be modified to clearly pick out the details in the periocular and ocular regions. For instance, Fig. 8(a) and Fig. 8(b) shows the before and after intensity transformation images respectively. Originally, the difference of black intensity values in the ocular region was much closer and hence, the details in the ocular region such as limbus (iris-sclera boundary) are not clear before the intensity transformation. In our research, we used the following two intensity transformation functions: (i) gamma transformation and (ii) contrast stretching.

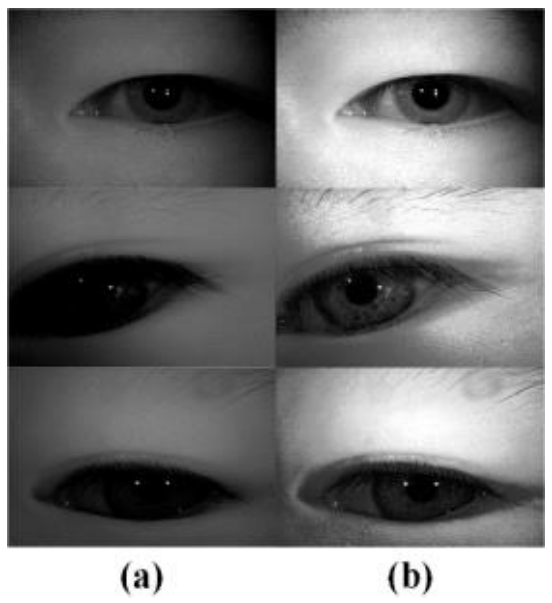

**Fig. 8: Intensity transformation (a) Original image (b) Transformed image.**

### 5.1.1.2.1 Gamma Transformation

Gamma transformation is mainly useful to enhance the bright (gray or white) information rooted in a large, predominantly dark region. The curve to transform the intensity values in an input image $\mathcal{X}$ to create the output image $\mathcal{D}$ is specified by the gamma parameter. If the value of gamma is greater than one, the intensities in an output image are weighted towards the darker side while, the values of gamma less than one, brighten the intensities in an output image. In our research, we omitted the gamma argument to create linear mapping of intensity values.

Additionally, four more parameters: low in $L_i$, high in $\mathcal{H}_i$, low out $L_o$ and high out $\mathcal{H}_o$ are used for clipping in gamma transformation. The intensity values in an input image $\mathcal{X}$ which are below the $L_i$ are clipped to $L_o$ and the intensity values greater than $\mathcal{H}_i$ are clipped to $\mathcal{H}_o$. We used the MATLAB function *stretchlim* to automatically compute the $L_i$ and $\mathcal{H}_i$ parameters and used the limit 0 and 1 for the $L_o$ and $\mathcal{H}_o$ respectively.

### 5.1.1.2.2 Contrast Stretching

Contrast stretching upsurges the contrast in an input image by changing its narrow range of intensities to span a desired wider range of output values. The contrast stretching transformation function has the form as expressed in Eq. (7):

$$\mathcal{D}(i,j) = T[\mathcal{X}(i,j)] = \frac{1}{1+(m/\mathcal{X}(i,j))^E} \tag{7}$$

Where $\mathcal{D}$ is the final denoised image in our proposed methodology. The resultant image after the gramma transformation $\mathcal{X}$ is fed as input to the contrast transformation function $T$. $m$ represents the mid-line value to switch between the light and dark values. We specified $m$ as the mean intensity value of input image $\mathcal{X}$. The slope of contrast stretching function is controlled by $E$, which in our case is equal to 3.

### 5.1.2 Image Enhancement

The image enhancement technique in our proposed framework is primarily based on contrast limited adaptive histogram equalization [96]. Firstly, we enhanced the dynamic range of denoised image $\mathcal{D}$ using histogram equalization. This step resulted in an image having pixel intensities evenly distributed across the entire range. Histogram equalization improved the contrast level especially in images where the intensities are clustered predominantly around the lower or middle range as shown in Fig. 9. We further enhanced this image using CLAHE technique to obtain the final enhanced image $\mathcal{E}$.

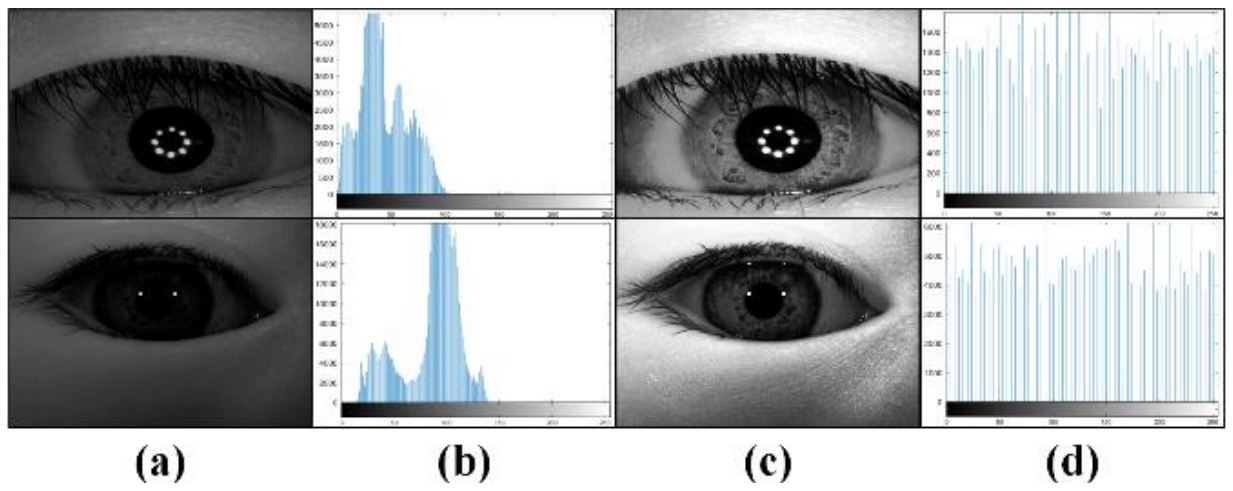

**Fig. 9: (a) Original image (b) Histogram of original image (c) Histogram equalization (d) Histogram of equalized image.**

In CLAHE technique, the small areas in an image called titles are processed instead of an entire image at once. Each title is enhanced individually based on the distribution and shape of the histogram selected. Finally, bilinear interpolation is used to merge the adjacent tiles and to abolish the artificially induced borders if any. In our research, we specified three parameters used in CLAHE as: (i) size of tile = 20×20, (ii) histogram shape = flat and (iii) limit of contrast enhancement = 0.005. Fig. 10(a), Fig. 10(b) and Fig. 10(c) shows the raw pristine images, final denoised images $\mathcal{D}$ and the enhanced images $\mathcal{E}$ for each of the five subsets of CASIA-IrisV4 database respectively. Significant enhancement can be observed in the iris region in Fig. 10(c), when compared with Fig. 10(a,b). This justifies the use of local enhancement technique (CLAHE) in addition to the global enhancement methods (gamma transformation and contrast stretching).

### 5.1.3 Reflection Removal

After enhancing the image, we removed two type of reflections present in the enhanced image during this stage: (i) reflection in the ocular region mainly due to cornea and aqueous humour and (ii) reflection in the periocular region mainly due to flashlights. The final filtered image after reflection removal is labeled as $\mathcal{F}$.

#### 5.1.3.1 Reflection Removal in Ocular Region

Our prime focus is to fill the holes in the ocular region (pupil, iris and sclera) while preserving the texture. In our study, using the simple hole filling strategy is not effective because it disrupts the texture of the image as shown in Fig. 11(c). This may lead to the false segmentation in the ocular region at later stages. Therefore, we adopted the strategy of adaptive thresholding [97] to first extract the holes as shown in Fig. 11(b). Then, we replaced these hole pixels with the corresponding filled hole pixels in the enhanced image as shown in Fig. 11(d). Adaptive thresholding scheme is an improved version as compared to the conventional algorithms, which are based on the global threshold value. It utilizes the dynamic threshold value for each pixel in the image, computed using local mean intensity in the pixel neighborhood.

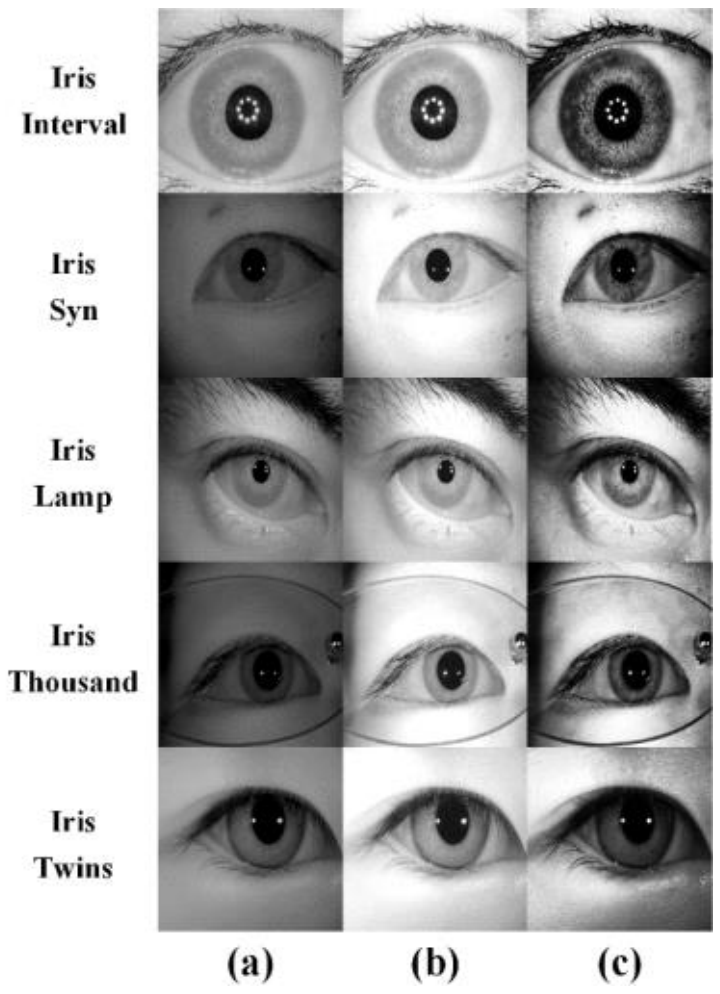

**Fig. 10: Image enhancement (a) Original image (b) Denoised image (c) Enhanced image.**

We specified the value of sensitivity as zero and set the foreground polarity as bright since the reflection in the ocular region appear as the pure bright holes. This resulted in the binarized image containing most of the ocular region holes. Afterwards, we applied the basic dilation operation on the extracted holes to enlarge their boundaries as shown in Fig. 11(b). We used the sphere shaped structuring element with radius of 2 pixels to perform the dilation operation.

#### 5.1.3.2 Reflection Removal in Periocular Region

Skin predominantly reflects the light in the periocular region and the similarity between intensities of the skin pixels is far more as compared to rest of the image. Based on these observations, we applied the non-local means (NLM) filtering technique [98] to remove the reflection in the periocular region.

NLM is an image denoising algorithm which differs from the local mean filter that computes the mean value in the target pixel neighborhood for smoothing. Whereas, NLM computes the mean of all similar pixels to the target pixel based on weighted Euclidean distance in the comparison window. The weight is a decreasing exponential function and another parameter called degree of smoothing, determines the rate of weight decay. NLM when compared with the local mean algorithms, produces far higher post-filtering clarity and much lesser loss of information in the image [98]. In our study, we specified the NLM parameters as: (i) degree of smoothing = 7, (ii) search window size = 25 and (iii) comparison window = 17. Fig. 11(e) shows the final filtered image $\mathcal{F}$ without the reflection.

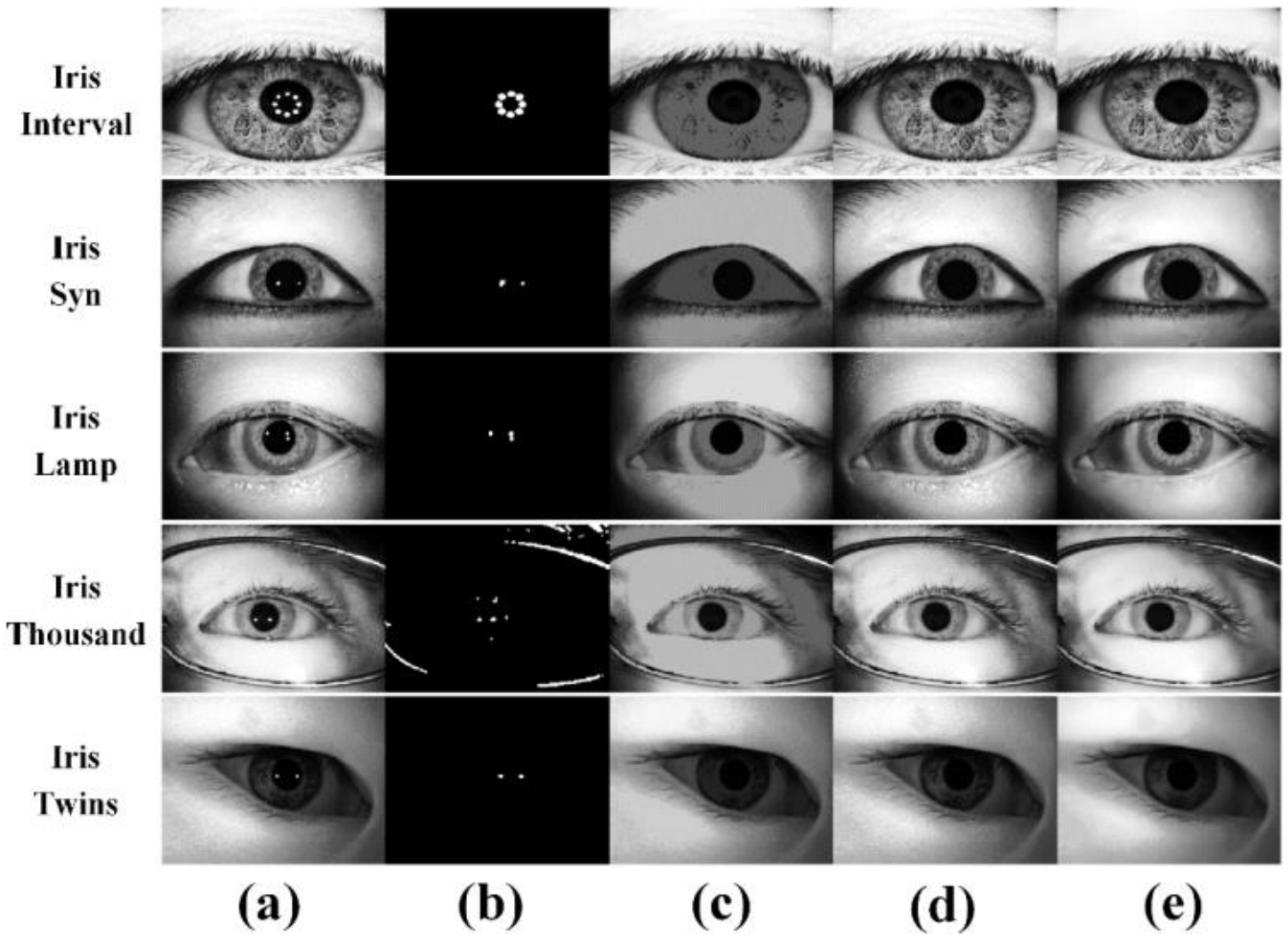

**Fig. 11: Reflection removal (a) Enhanced image, (b) Extraction of bright reflection holes, (c) Reflection removal using simple hole filling strategy, (d) Reflection removal using adaptive thresholding strategy (e) Reflection removal in periocular region using NLM filtering.**

### 5.1.4 Extraction of Periocular Components

The core components in the periocular region as evident from Fig. 1 are eyebrows, wrinkles, spectacles, upper/ lower eyelids and eyelashes. Usually, there is a sharp transition at the boundary of these components and hence they can be referred to as high frequency components. The aim is to extract these high frequency components in the periocular region and attenuate them, which we achieved in the next stage using fuzzy filters.

In our study, extraction of the periocular components is a two phase process. At first, we binarized the final enhanced image $\mathcal{E}$ using adaptive thresholding technique by setting the parameters of adaptive thresholding as: (i) sensitivity = 0.375, and (ii) foreground polarity = dark. This concludes the first phase, which resulted in the image $\mathcal{P}'$ containing periocular components as shown in Fig. 12(b). We used the adaptive thresholding approach to extract the periocular components as it is more flexible and can adapt to the changing lighting conditions in the image, for instance, strong illumination gradient or shadows.

Since the resultant image $\mathcal{P}'$ also contains the pupil as shown in Fig. 12(b), which is not meant to be attenuated, therefore, we have removed the pupil region using Daugman's integro differential operator (DIO) algorithm [33] as defined in Eq. (8):

$$max_{(r,i_0,j_0)} = \left| \mathcal{G}_\sigma(r) * \frac{\partial}{\partial_r} \oint_{r,i_0,j_0} \frac{\mathcal{P}\prime(i,j)}{2\pi r} ds \right| \quad (8)$$

Where $\mathcal{P}'(i,j)$ is the intensity value of pixel at location $(i,j)$ and $\mathcal{G}$ is the Gaussian smoothing filter with standard deviation $\sigma$. $s$ is the contour of circle with center coordinates $i_0, j_0$ and the radius $r$. DIO algorithm is an iterative process which locates the circular path based on maximum change in the pixel intensity value, by varying the contour parameters $r, i_0, j_0$.

Daugman's operator in our study is based on the observation that difference between the pixel intensity values at the contour of pupil is greater than any other circle in the $\mathcal{P}'$ as shown in Fig. 12(c). We limited the DIO algorithm to discard the false pupil contours by specifying 5 as the minimum radius size. After obtaining the boundary of pupil region, we filled zeros in it to create a pupil mask as shown in Fig. 12(d). We obtained the final periocular region $\mathcal{P}$ in our study by subtracting the pupil mask from the $\mathcal{P}'$ as shown in Fig. 12(e).

### 5.1.5 Blurring of Periocular Components

This is the final step in our preprocessing stage and the aim of this step is to suppress the periocular information in the enhanced filtered image $\mathcal{F}$. For this, we used the asymmetrical triangular fuzzy filter with median center (ATMED) [99]. Fuzzy filters are often used for smoothing in the domain of image processing. Eq. (9) defines the output $\mathfrak{F}(i,j)$ of the general 2-dimensional fuzzy filter.

$$\mathfrak{F}(i,j) = \frac{\sum_{(r,s)\in\mathcal{A}} \mathcal{W}[\mathcal{F}(i+r,j+s)] * \mathcal{F}(i+r,j+s)}{\sum_{(r,s)\in\mathcal{A}} \mathcal{W}[\mathcal{F}(i+r,j+s)]} \tag{9}$$

Where, $\mathcal{F}$ is the input image, $\mathcal{W}[\mathcal{F}(i,j)]$ is the window function with area $\mathcal{A}$ at location $(i,j)$. The image pixel in window $\mathcal{A}$ is represented by $(r,s) \in \mathcal{A}$. The window function for ATMED $\mathcal{W}_{ATMED}[\mathcal{F}(i+r,j+s)]$ is defined using Eq. (10):

$$\mathcal{W}_{ATMED}[\mathcal{F}(i+r,j+s)] = \begin{cases} 1 - \frac{\mathcal{F}_{med}(i,j) - \mathcal{F}(i+r,j+s)}{\mathcal{F}_{med}(i,j) - \mathcal{F}_{min}(i,j)} \\ for\ \mathcal{F}_{min}(i,j) \le \mathcal{F}(i+r,j+s) \le \mathcal{F}_{med}(i,j) \\ 1 - \frac{\mathcal{F}(i+r,j+s) - \mathcal{F}_{med}(i,j)}{\mathcal{F}_{max}(i,j) - \mathcal{F}_{med}(i,j)} \\ for\ \mathcal{F}_{med}(i,j) \le \mathcal{F}(i+r,j+s) \le \mathcal{F}_{max}(i,j) \\ 1 \\ for\ \mathcal{F}_{med}(i,j) - \mathcal{F}_{min}(i,j) = 0 \\ or\ \mathcal{F}_{max}(i,j) - \mathcal{F}_{med}(i,j) = 0 \end{cases} \tag{10}$$

Where, the difference between $\mathcal{F}_{med}(i,j) - \mathcal{F}_{min}(i,j)$ and $\mathcal{F}_{max}(i,j) - \mathcal{F}_{med}(i,j)$ determines the degree of asymmetry. $\mathcal{F}_{max}(i,j)$, $\mathcal{F}_{min}(i,j)$ and $\mathcal{F}_{med}(i,j)$ are the maximum, minimum and median values of all the input values $\mathcal{F}(i+r,j+s)$ for $(r,s) \in \mathcal{A}$ within the window $\mathcal{A}$ at discrete indexes $(i,j)$, respectively.

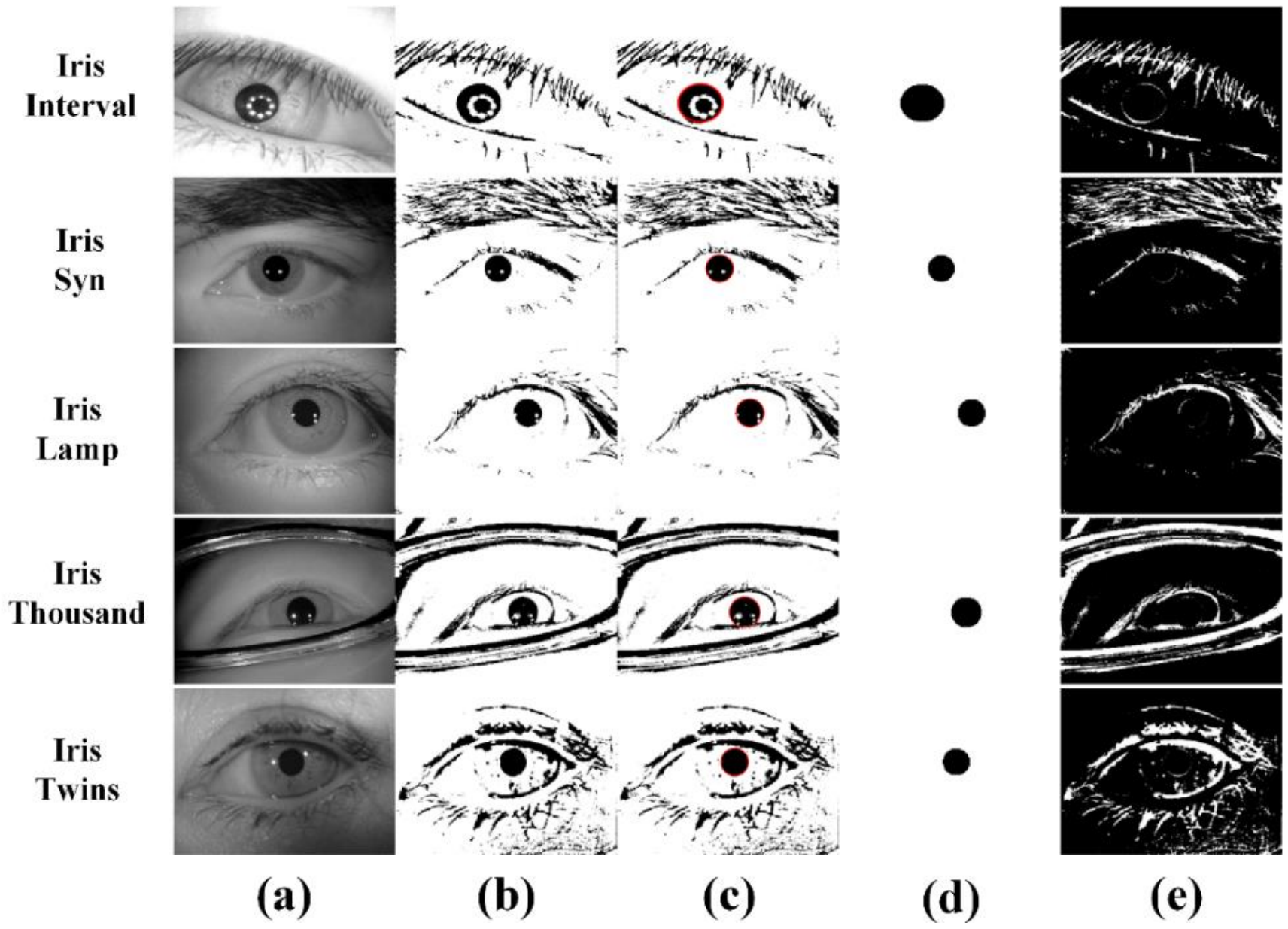

**Fig. 12: Extraction of periocular components (a) Original image, (b) Extraction of periocular components with pupil (c) Detection of pupil contour using DIO algorithm (shown in red color boundary), (d) Pupil mask generation, (e) Final periocular region after subtraction of pupil.**

In our research, the filtered image $\mathcal{F}$ obtained during the reflection removal step is first passed to the ATMED fuzzy function. We specified the local search window size $W_s$ = 21 and the padding value for window $P_w$ is determined using $P_w = (W_s - 1)/2$. This resulted in the fuzzified blur image $\mathfrak{F}$ as shown in Fig. 13(b).

Later, using the filtered image $\mathcal{F}$, the fuzzified blur image $\mathfrak{F}$ and the final periocular image $\mathcal{P}$, we suppressed the periocular components (where $\mathcal{P}$ = 1), by replacing these pixels in the filtered image $\mathcal{F}$ with the corresponding intensity values of fuzzified blur image $\mathfrak{F}$. This produces the final preprocessed image $\mathfrak{P}$ as shown in Fig. 13(d), which is later fed to the CNN for segmentation. The working flow of our preprocessing stage is summarized in Algorithm I and Fig. 14 shows the randomly selected preprocessing results from each of the five datasets used in this study.

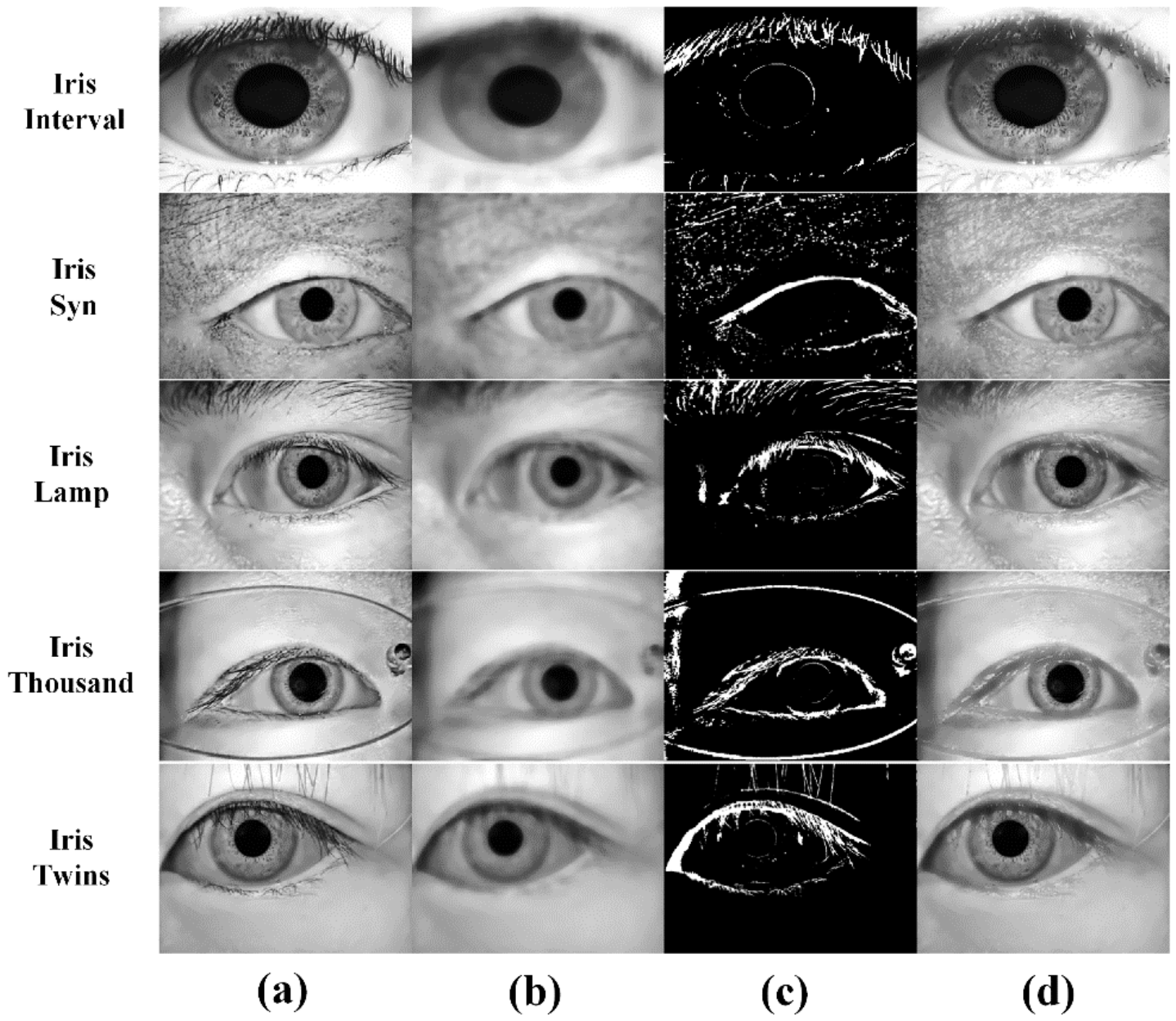

**Fig. 13: Periocular suppression (a) Filtered image (b) Fuzzified image (c) Periocular region (d) Periocular suppression.**

**ALGORITHM I: Preprocessing Framework**

<START>
$\mathcal{X} \leftarrow$ Load the Input Image
1: $[\mathcal{D}]$ = Image Denoising $(\mathcal{X})$
$\mathcal{D}_1$= Denoise $\mathcal{X}$ using DnCNN
$\mathcal{D}$ = Intensity Transformations on $\mathcal{D}_1$
2: $[\mathcal{E}]$ = Image Enhancement $(\mathcal{D})$
$\mathcal{E}_1$ = Histogram Equalization on $\mathcal{D}$
$\mathcal{E}$ = CLAHE on $\mathcal{E}_1$
3: $[\mathcal{F}]$ = Reflection Removal $(\mathcal{E})$
$\mathcal{F}_1$ = Adaptive Thresholding on $\mathcal{E}$
$\mathcal{F}_2$ = Binarize $\mathcal{E}$ using $\mathcal{F}_1$
$\mathcal{F}_3$ = Fill Holes in $\mathcal{E}$
$\mathcal{F}_4$ = Replace Pixels in $\mathcal{E}$ at $\mathcal{F}_2$==1 with Corresponding Pixels in $\mathcal{F}_3$
$\mathcal{F}$ = NLM Filtering on $\mathcal{F}_4$
4: $[\mathcal{P}]$ = Periocular Region Extraction $(\mathcal{E})$
$\mathcal{P}_1$ = Adaptive Thresholding on $\mathcal{E}$
$\mathcal{P}'$ = Binarize $\mathcal{E}$ using $\mathcal{P}_1$
$\mathcal{P}_2$ = Detect Pupil using DIO Algorithm
$\mathcal{P}_3$ = Create Pupil Mask
$\mathcal{P} = \mathcal{P}' - \mathcal{P}_3$
5: $[\mathfrak{P}]$ = Periocular Suppression $(\mathcal{F}, \mathcal{P})$
$\mathfrak{F}$ = Fuzzy Filtering on $\mathcal{F}$
$\mathfrak{P}$ = Replace Pixels in $\mathcal{F}$ at $\mathcal{P}$ ==1 with Corresponding Pixels in $\mathfrak{F}$
Final Preprocessed Image $\rightarrow \mathfrak{P}$
<END>

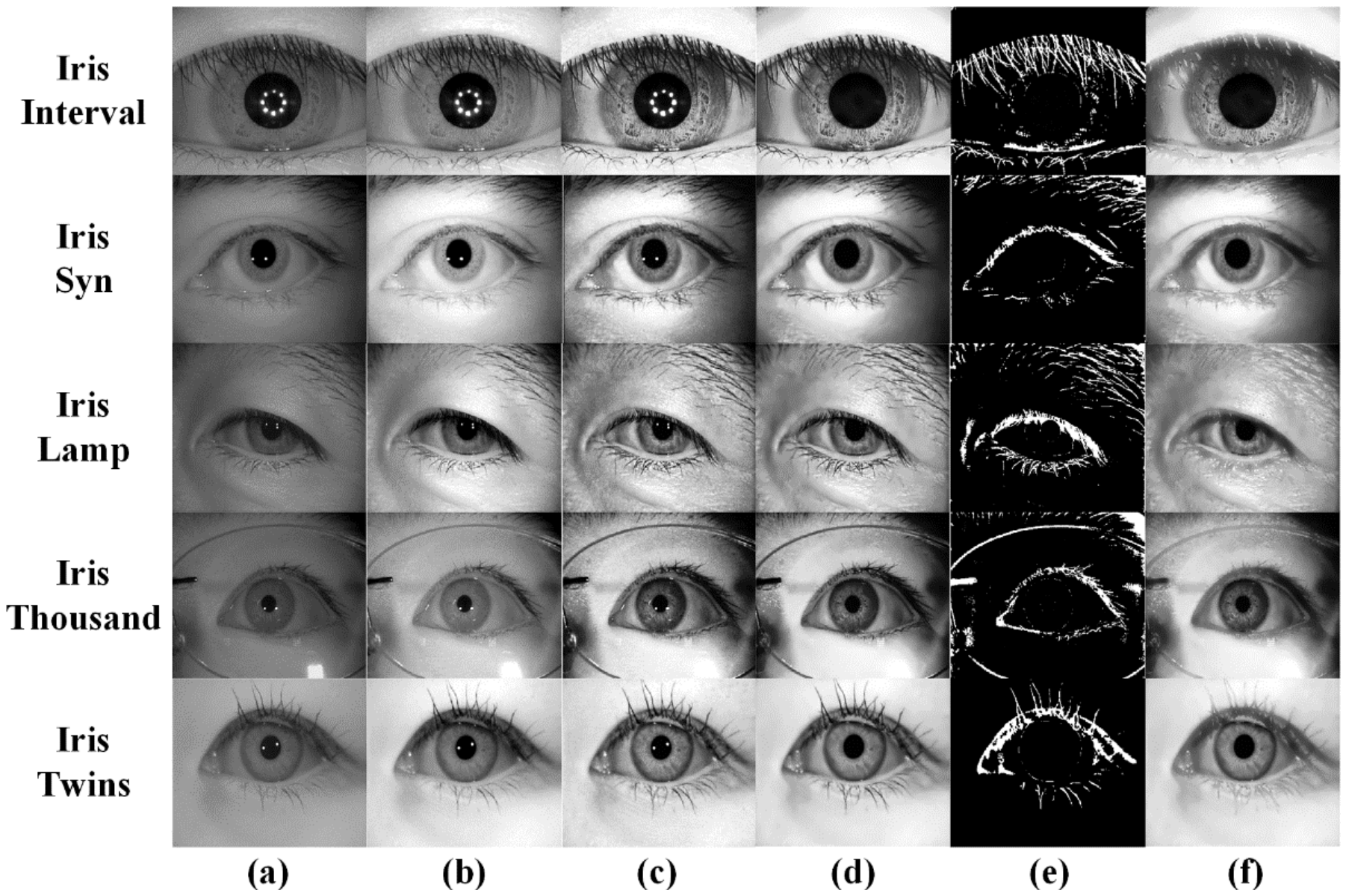

**Fig. 14: Preprocessing stage (a) Original image (b) Denoised image (c) Enhanced image (d) Image with reflection removed (e) Periocular region (f) Preprocessed image with periocular suppression.**

## 5.2 Image Denoising

We randomly divided each of the five subsets of CASIA-IrisV4 database in the ratio of 60:20:20 for training, validation and testing purpose respectively. SIP-SegNet is trained using 60% of the images in each subset, while for both validation and testing, we used 20% of the images in each subset. Table IX shows the details of training, validation, and test sets used in this study.

**Table IX: Details of Training, Validation and Test Sets.**

| Subsets | Total Images | Training (60%) | Validation (20%) | Test (20%) |
|---|---|---|---|---|
| **CII** | 2639 | 1583 | 528 | 528 |
| **CIS** | 10000 | 6000 | 2000 | 2000 |
| **CIL** | 16212 | 9728 | 3242 | 3242 |
| **CITW** | 3183 | 1909 | 637 | 637 |
| **CIT** | 20000 | 12000 | 4000 | 4000 |
| **Total** | **52034** | **31220** | **10407** | **10407** |

## 5.3 Data Augmentation

Data augmentation can alleviate overfitting, which mainly results from training a network on a small dataset [100]. Various image augmentation techniques such as cropping, rotation, translation etc. produce new distinct images and thus provide a mean of leveraging limited training data. Data augmentation may also improve the accuracy of the network, since it provides more diversified training data to the network. We applied the following augmentations to the training data in our research.

- **Reflection –** We used random reflection in both X (left-right) and Y (top-bottom) dimensions. This transformation flipped the training images with 50% probability in each dimension.
- **Rotation –** We specified the range of rotation as [-30° 30°]. This transformation rotated the training images by randomly picking a rotation angle within the specified interval.
- **Scaling –** We set the scaling range as [1.2 1.5] times. This transformation scaled the training images by randomly selecting a scale factor from the range. The images are resized by the scale factor in both the horizontal and vertical directions.

- **Translation –** The translation range used is [-20 20] pixels. This transformation shifted the training images both horizontally and vertically by selecting a distance from the range randomly.

We trained the network using augmented datastore, which contains both the training images and augmentation parameters. This augmented datastore perturbs the training data for each epoch by applying random transformations as shown in Fig. 15. However, it does not change the actual number of training images in each epoch.

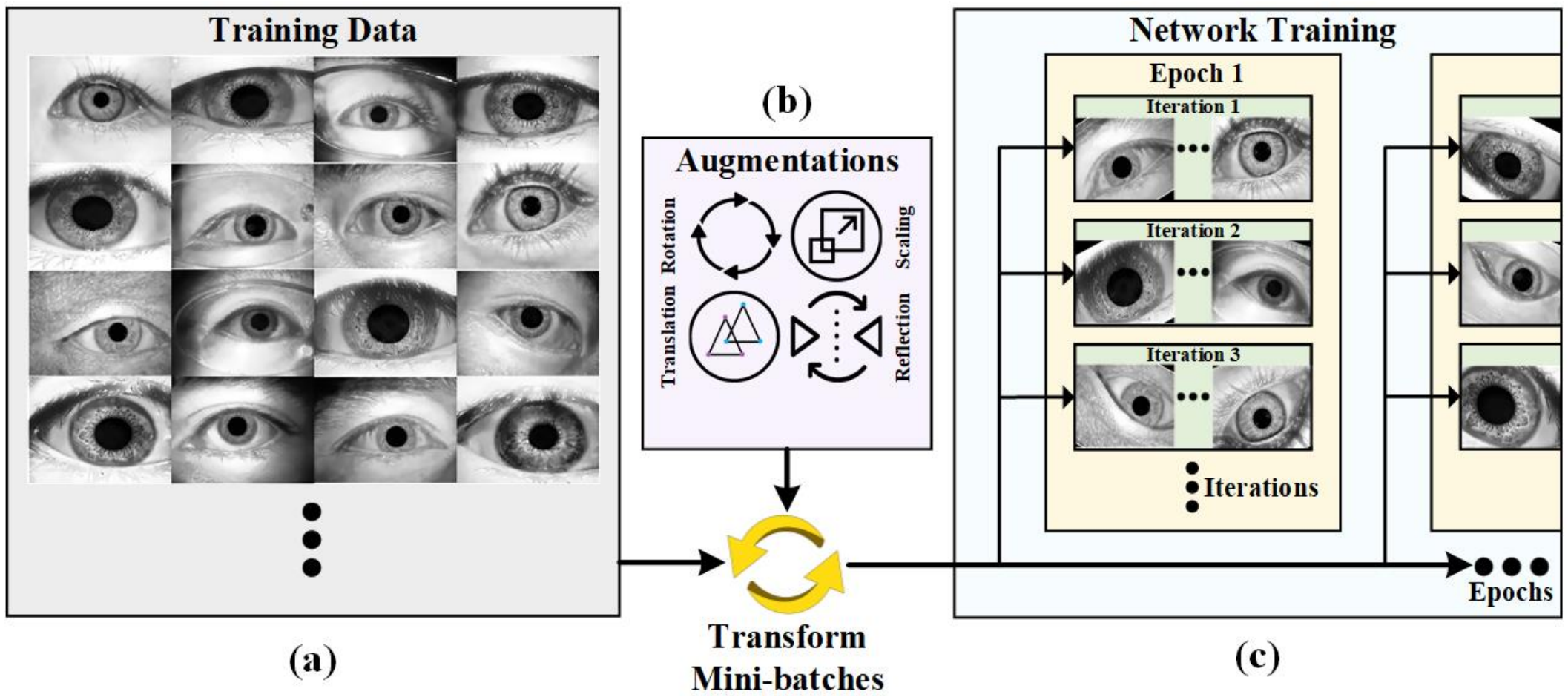

**Fig. 15: Augmentation process (a) Training data consisting of preprocessed images (b) Various image augmentation techniques: rotation, scaling, translation and reflection (c) Depiction of network training using augmented data.**

## 5.4 SIP-SegNet Architecture

Our proposed architecture for semantic segmentation of iris, pupil and sclera (SIP-SegNet) uses the pretrained encoder-decoder based framework called Segnet [32] by fine tuning some of its parameters. Encoder-decoder implementation is based on the understanding that CNN models via layer-to-layer propagation can automatically extract the high-level features in the spatial domain. However, the downsampling operation results in the loss of spatial information and to regain this information, we need to perform the symmetrical upsampling or deconvolution.

Fig. 16(b) shows the architecture of SIP-SegNet, where the encoder path comprises of five encoders. The topology of encoder path in SIP-SegNet is same as that of VGG [101] but without the fully connected layers. The VGG-16 network shown in the Fig. 16(a) is used as a backbone of SIP-SegNet in this research. The encoder path in SIP-SegNet consists of 13 convolutional layers in total and is used to generate a set of feature maps by performing a convolution with a 3×3 filter banks. The first two encoders contain two convolutional layers each whereas, the next three encoders consist of three convolutional layers each. Also, while configuring all the convolutional layers in the proposed network, the bias term is fixed to zero. Batch normalization (BN) and element-wise rectified linear unit (ReLU) operations are then applied to these set of feature maps. In addition, each of the five encoders are followed by the max-pooling operation with a 2×2 window and stride 2. The spatial resolution of the feature maps is reduced by a factor of 2 after each successive pooling operation. The size of the feature map at the output of the encoder network is 7×7, which is 1/32 times the size of the input image (224×224).

The reduced representation (7×7) of an image is not suitable for performing the segmentation task, where the contour delineation is essential. Therefore, the encoder network in SIP-SegNet is followed by the decoder network, which is responsible to increase the spatial resolution of feature maps in order to match with the dimensions of the input image in a stepwise manner. The SIP-SegNet decoder network is inspired from an unsupervised feature learning architecture [102] consisting of hierarchy of decoders i.e. each decoder corresponding to each encoder. Each decoder performs the non-linear upsampling of the input feature maps by using the corresponding encoder's max-pooling indices and trainable filters as shown in Fig. 16(b) and Fig. 17. The decoder completes the pixel-wise classification using multi-class soft-max classifier. The soft-max classifier layer outputs a K channel image, where K = 4 is the total number of desired classes, having probability value against each pixel. One of the key features in SIP-SegNet framework is the direct transfer of information instead of convolving it. Segnet [32] is considered as one of the leading models especially when dealing with the image segmentation problems. Table X shows the architectural details of SIP-SegNet model used in this study.

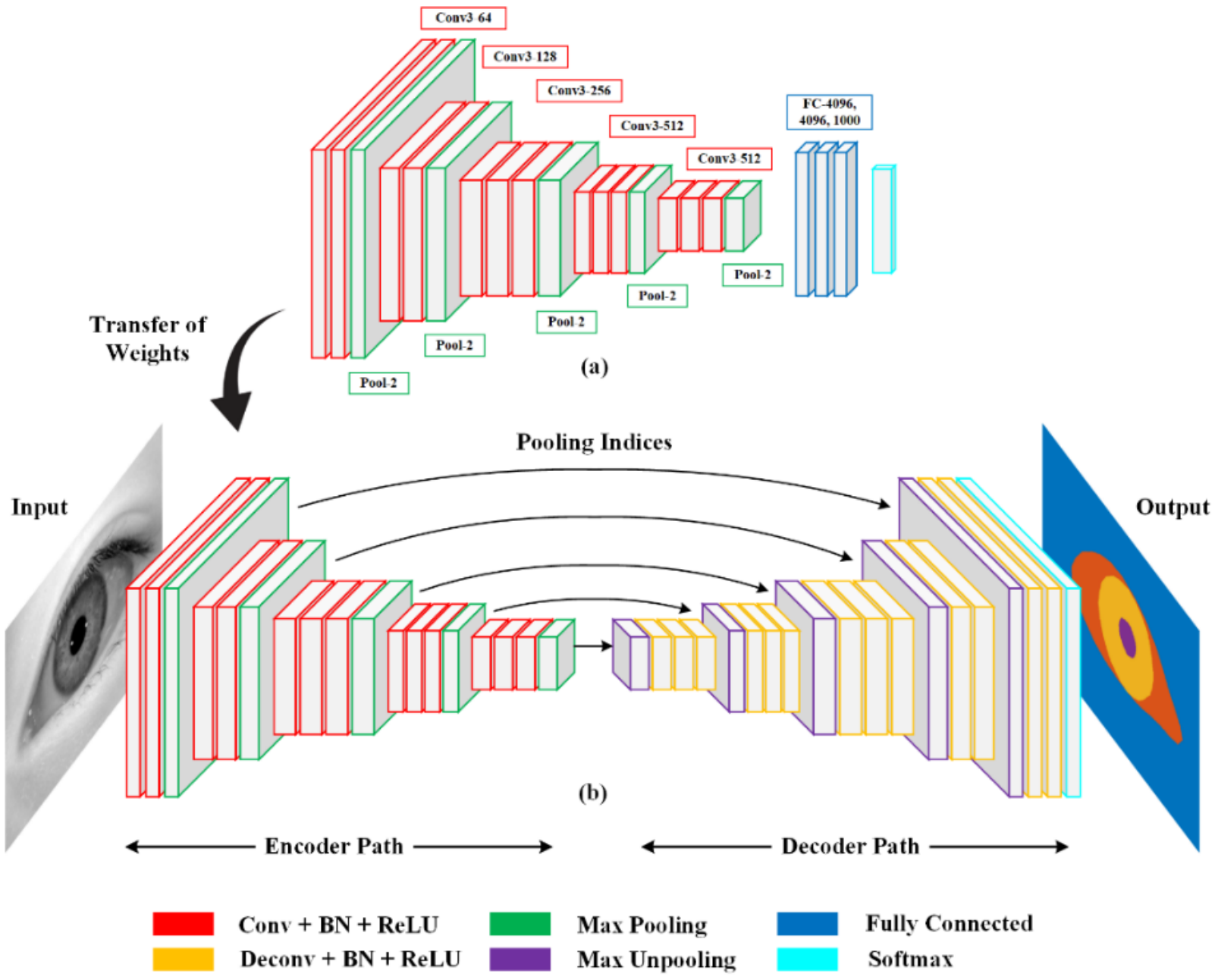

**Fig. 16: (a) VGG-16 architecture, (b) SIP-SegNet architecture.**

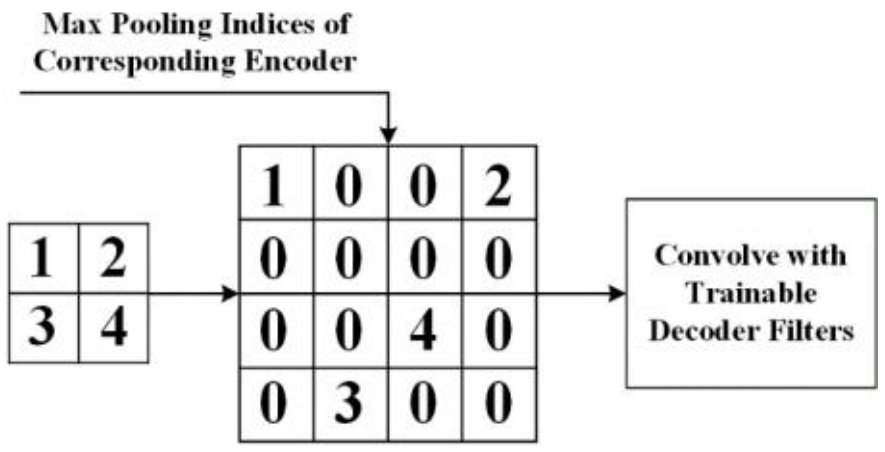

**Fig. 17: SIP-SegNet decoder module.**

**Table X: SIP-SegNet Architecture.**

| Set | Layers | Maps | Kernel | Stride | Output Size |
|---|---|---|---|---|---|
| Input | Image | 1 | - | - | 224×224×3 |
| E-1 | (C+BN+R) ×2 | 64 | 3×3 | 1 | 224×224×64 |
| | MP | | | 2 | 112×112×64 |
| E-2 | (C+BN+R) ×2 | 128 | 3×3 | 1 | 112×112×128 |
| | MP | | | 2 | 56×56×128 |
| E-3 | (C+BN+R) ×3 | 256 | 3×3 | 1 | 56×56×256 |
| | MP | | | 2 | 28×28×256 |
| E-4 | (C+BN+R) ×3 | 512 | 3×3 | 1 | 28×28×512 |
| | MP | | | 2 | 14×14×512 |
| E-5 | (C+BN+R) ×3 | 512 | 3×3 | 1 | 14×14×512 |
| | MP | | | 2 | 7×7×512 |
| D-5 | MUP | 512 | 3×3 | 2 | 14×14×512 |
| | (DC+BN+R) ×3 | | | 1 | 14×14×512 |
| D-4 | MUP | 512 | 3×3 | 2 | 28×28×512 |
| | (DC+BN+R) ×3 | | | 1 | 28×28×512 |
| D-3 | MUP | 256 | 3×3 | 2 | 56×56×256 |
| | (DC+BN+R) ×3 | | | 1 | 56×56×256 |
| D-2 | MUP | 128 | 3×3 | 2 | 112×112×128 |
| | (DC+BN+R) ×2 | | | 1 | 112×112×128 |
| D-1 | MUP | 64 | 3×3 | 2 | 224×224×64 |
| | (DC+BN+R) ×2 | | | 1 | 224×224×64 |
| PC | Softmax | 4 | - | - | 224×224×4 |

E= Encoder, D= Decoder, C= Convolutional Layer, BN= Batch Normalization Layer, R= Rectified Linear Unit, MP= Max Pooling, DC= Deconvolutional Layer, MUP= Max Unpooling, PC= Pixels Classification Layer.

## 5.5 Class Balancing

Ideally, the pixel frequency of all classes in the training data should be equal. However, the training pixel frequencies for the four classes (periocular, sclera, iris and pupil) in our dataset are highly imbalanced. The pixels belonging to the periocular class are very large in number as compared to the other three classes. This is because the periocular region covers the most area in the ocular dataset used in this study. Fig. 18 shows the statistical distribution of the pixel labels of each class. Subsequently, there is a need to resolve the imbalance between the four classes as evident from the Fig. 18, because the network learning process is biased towards the dominant class and such imbalance can affect the learning process negatively.

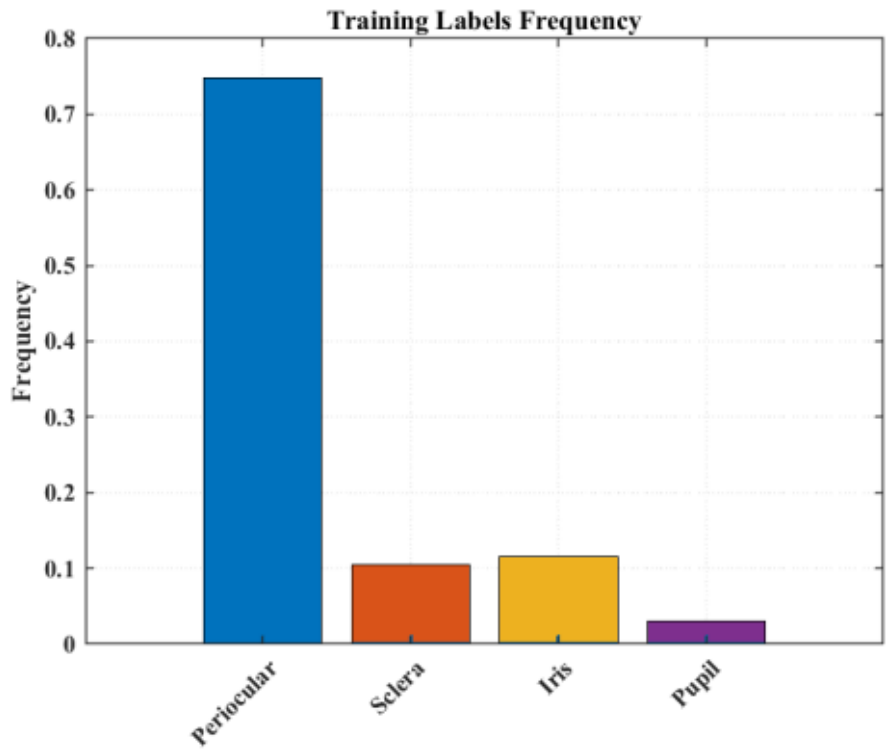

**Fig. 18: Frequency of training pixel labels.**

We used the inverse frequency weighting approach to balance out the underrepresented classes in the training data. This approach computes the class weights by taking the inverse of each class frequency as expressed in Eq. (11,12). These class weights are specified in the last layer of the SIP-SegNet model to cater the imbalance between the classes.

$$Frequency = \frac{Pixels(i)}{Total\ Pixels} \tag{11}$$

$$Class\ Weights = \frac{1}{Frequency} \tag{12}$$

Where, $Pixels(i)$ is the total number of pixels belonging to class $i$ in the training data. In our proposed study, $i$ = 4 represents the four classes: periocular, sclera, iris and pupil.

## 5.6 Training Parameters

The loss function used during the training phase of this study is mean binary cross entropy between the output and the target as expressed in Eq. (13):

$$\mathcal{L} = \frac{1}{M \times N \times B} \sum_{c=1}^{B} \sum_{b=1}^{N} \sum_{a=1}^{M} t_{ab}\, log(p_{ab}) - (1 - t_{ab}) log(1 - p_{ab}) \tag{13}$$

Where, $t_{ab}$ and $p_{ab}$ represent the values of pixel at location $(a, b)$ in the target image and the output image respectively. $M \times N$ shows the dimension of an image and $B$ is the batch size. We used SGDM as the optimization method. Adding the momentum parameter boosts the convergence and also prevents the gradient descent to oscillate along the local minimums. This defines the contribution of the gradient decent from the previous iteration to the current iteration. Consider $\mathcal{L}(\omega)$ as the loss function with network parameters $\omega$, then SGDM method is defined using Eq. (14):

$$\omega := \omega - \eta \nabla \mathcal{L}(\omega) + \alpha \Delta \omega \tag{14}$$

Here $\eta$ and $\alpha$ represent the learning rate and momentum respectively. $\Delta\omega$ is the update in the previous iteration and $\nabla \mathcal{L}(\omega)$ is the gradient value in the current iteration. The hyper parameters of the network are empirically fine-tuned as listed in the Table XI. The training performance of SIP-SegNet is shown in Fig. 19. We specified the piecewise schedule for learning rate and reduced it by a factor of 0.3 at every 40 epochs. As a result, a higher initial learning rate allowed the network to learn quickly, and once the learning rate drops it can find a solution close to the local optimum.

Since, the training data is very large, we specified the mini-batch size of 32 to define the subset of training dataset used to update the weights and the loss function gradient. The reason for specifying a small mini-batch size is to reduce the memory usage because semantic segmentation consumes a lot of memory. The training data is shuffled before each epoch, while the validation data is shuffled before each network validation.

To prevent the network from overfitting, we used the regularization term (weigh decay) and validation patience. Validation patience stops the training early if the validation accuracy converges. We set the validation patience = 4, which allowed the loss on the validation data to exceed the previously smallest loss only four times before the network training stops.

**Table XI: SIP-SegNet Training Hyper Parameters**

| Parameters | Value |
|---|---|
| Training Images | 31220 |
| Validation Images | 10407 |
| Maximum Epochs | 120 |
| Mini-Batch Size | 32 |
| Iterations/ Epoch | 976 |
| Total Iterations | 117120 |
| Validation Frequency | 7150 |
| Solver | SGDM |
| Momentum | 0.9 |
| Learning Rate | 0.01 |
| Gradient Threshold | 0.005 |
| Weight Decay | 0.0005 |

SGDM= Stochastic Gradient Descent with Momentum.

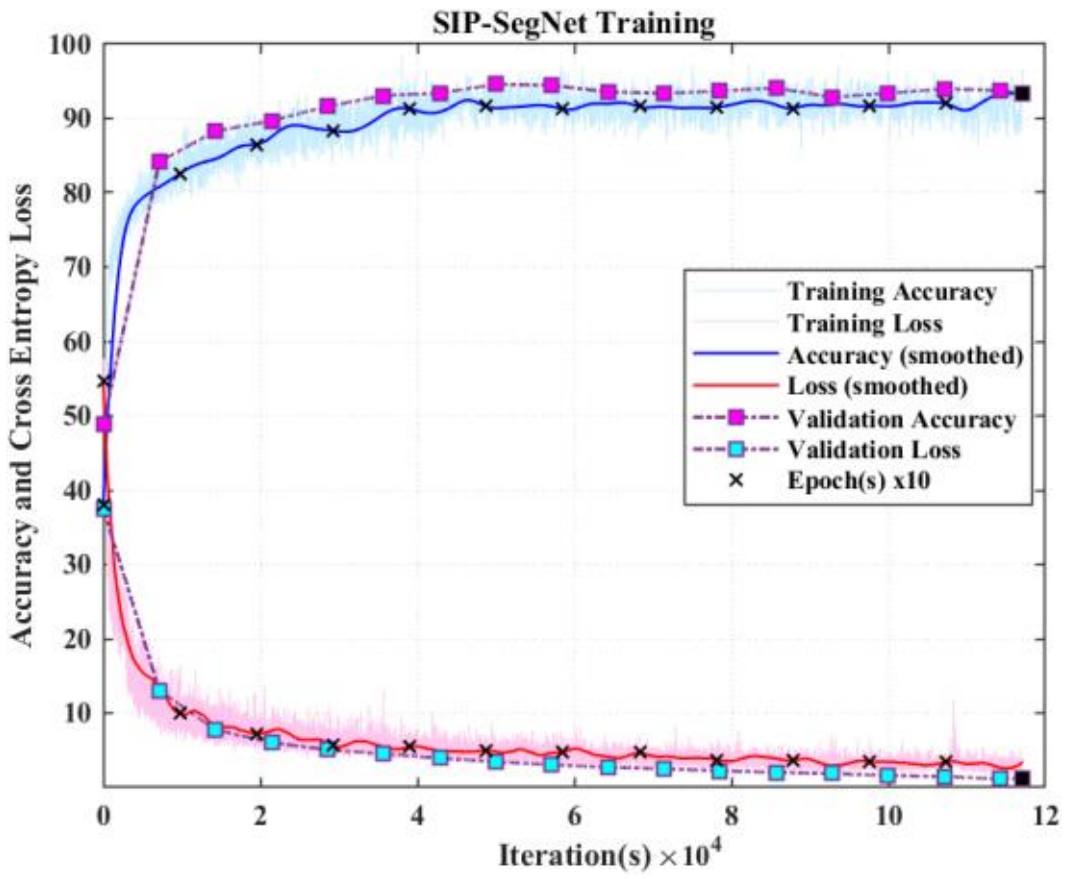

**Fig. 19: SIP-SegNet training performance.**

# 6. Experimental Setup

## 6.1 System Specifications

We used MATLAB R2019a simulation platform running on 64bits windows OS. The machine is configured as Intel Core i9-9980XE @3.0GHz, with 16GB memory and Nvidia GeForce RTX 2080.

## 6.2 Evaluation Metrics

In our research, we have used the following evaluation metrics to validate the performance of SIP-SegNet:

### 6.2.1 Confusion Matrix

Let $N_c$ be the total number of classes, $N_{xx}$ be the number of pixels that belong to class $x$ and predicted as $x$, then confusion matrix can be defined as a square table having number of true positives $N_{xx}$, true negatives $N_{yy}$, false positives $N_{xy}$ and false negatives $N_{yx}$. Fig. 20 illustrates the methodology for computing $N_{xx}$, $N_{yy}$, $N_{xy}$ and $N_{yx}$ in the semantic segmentation scenario, where the true class pupil (yellow circle) is exemplified.

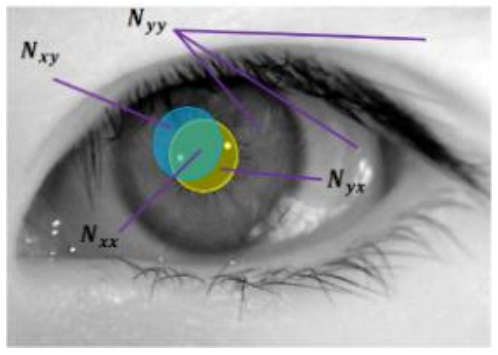

**Fig. 20: Derived terminologies from confusion matrix in semantic segmentation scenario. Yellow and blue circles represent the ground truth (pupil) and the predicted (pupil) labels respectively. $N_{xx}$ = true positive pixels, $N_{yy}$ = true negative pixels, $N_{xy}$ = false positive pixels and $N_{yx}$ = false negative pixels.**

### 6.2.2 Accuracy

Accuracy is the ratio of correctly classified labels to the total number of labels for each class in the dataset. It is computed as expressed in Eq. (15):

$$A = \frac{N_{xx}+N_{yy}}{N_{xx}+N_{yy}+N_{xy}+N_{yx}} \tag{15}$$

### 6.2.3 Precision

Precision is also known as positive predictive value, which shows the purity of true positives relative to the ground truth. Precision ($P$) for each class in a single image is computed using Eq. (16):

$$P = \frac{N_{xx}}{N_{xx}+N_{xy}} \tag{16}$$

### 6.2.4 Recall

This metric is also known as sensitivity or true positive rate, which shows the completeness of true positives relative to the ground truth. Recall ($R$) for each class in a single image is computed as expressed in Eq. (17):

$$R = \frac{N_{xx}}{N_{xx}+N_{yx}} \tag{17}$$

### 6.2.5 Specificity

Specificity also known as true negative rate or selectivity, is the model's ability to recall true negatives over all the negative samples. Specificity ($S$) for each class in a single image is computed as expressed in Eq. (18):

$$S = \frac{N_{yy}}{N_{xy}+N_{yy}} \tag{18}$$

### 6.2.6 Negative Predictive Value

Negative predictive value ($NPV$) shows the fraction of negative outcomes that are true negatives. Eq. (19) defines the $NPV$ for each class in a single image:

$$NPV = \frac{N_{yy}}{N_{yy}+N_{yx}} \tag{19}$$

### 6.2.7 Mean Accuracy

This metric shows the performance of each class in terms of percentage for correct classification of pixels. It is computed as expressed in Eq. (20):

$$I = \frac{1}{N_c}\left[\sum_{i=1}^{Nc}\left(\frac{N_{xx}(i)}{N_{xx}(i)+N_{yx}(i)}\right)\right] \tag{20}$$

The expression in Eq. (21) computes the mean average of all classes in single image $I$. Using this, the mean accuracy of complete dataset is computed as:

$$MA = \frac{1}{T}\left[\sum_{t=1}^{T}(I(i))\right] \tag{21}$$

Where, $MA$ is the mean accuracy of aggregate data, $T$ is the total number of images and $I$ is the mean accuracy of single image.

### 6.2.8 Global Accuracy

Global accuracy shows the overall performance in terms of correctly classified pixels. It is computed using the Eq. (22):

$$GA = \frac{1}{I}\left[\sum_{i=1}^{I}\left(\frac{N_{xx}(i)+N_{yy}(i)}{N_{xx}(i)+N_{yy}(i)+N_{xy}(i)+N_{yx}(i)}\right)\right] \quad (22)$$

Where $GA$ is the global accuracy and $I$ is the total number of images in the dataset. $GA$ shows the accuracy regardless of the class.

### 6.2.9 Mean Intersection over Union

Mean intersection over union (*MIoU*) is also called Jaccard similarity coefficient. It is the ratio between intersection and union, where for each class intersection is the number of true positives and union is the sum of true positives, false negatives and false positives. *MIoU* for all classes in a single image is computed as expressed in Eq. (23):

$$MIoU = \frac{1}{N_c}\left[\sum_{i=1}^{Nc}\left(\frac{N_{xx}(i)}{N_{xx}(i)+N_{xy}(i)+N_{yx}(i)}\right)\right] \quad (23)$$

### 6.2.10 Frequency Weighted Intersection over Union

Frequency weighted intersection over union (*FWIoU*) weighs each class *IoU* by the total number of pixels in it. It is a useful metric where the class sizes are disproportional. In such cases, the impact of error in the small classes is reduced by aggregate quality score. *FWIoU* is computed using Eq. (24):

$$FWIoU = \left(\sum_{m=1}^{N_c} P_m\right)^{-1}\left[\sum_{i=1}^{Nc}\left(\frac{N_{xx}(i)}{N_{xx}(i)+N_{xy}(i)+N_{yx}(i)}\right)\right] \quad (24)$$

Where $P_m$ is the total number of pixels in each class.

### 6.2.11 Dice Similarity Coefficient

Dice similarity coefficient is used to measure the degree of similarity between the predicted and ground truth image. Dice for all classes in a single image is computed as expressed in Eq. (25):

$$D = \frac{1}{N_c}\left[\sum_{i=1}^{Nc}\left(\frac{2\times N_{xx}(i)}{(2\times N_{xx}(i))+N_{xy}(i)+N_{yx}(i)}\right)\right] \quad (25)$$

### 6.2.12 Boundary F1 Score

Boundary F1 score is also known as contour matching score and it shows how closely the predicted contour of each class aligns with the ground truth contour. F1 score in terms of precision and recall for each class in a single image is computed as expressed in Eq. (26):

$$F1 = 2\left[\frac{P\times R}{P+R}\right] \quad (26)$$

### 6.2.13 Nice1

Nice1 is the segmentation error score. It computes the fraction of corresponding disagreeing pixels in an image using Eq. (27):

$$N1 = \frac{1}{M\times N\times T}\left[\sum_{k=1}^{T}\sum_{i,j\in(M,N)} G(i,j)\oplus P(i,j)\right] \quad (27)$$

Here $T$ represents the total number of images with spatial resolution $M \times N$. $G(i,j)$ and $P(i,j)$ are the pixels of ground truth and the predicted labels respectively.

### 6.2.14 Nice2

Nice2 is another metric to represent an error score. It is computed by taking the average of the sum between the false positive rate (FPR) and false negative rate (FNR) as expressed in Eq. (28):

$$N2 = \frac{1}{2}[FPR + FNR] \quad (28)$$

Where FPR and FNR for each class in a single image are defined as expressed in Eq. (29,30):

$$FPR = \frac{N_{xy}}{N_{xy}+N_{yy}} \tag{29}$$

$$FNR = \frac{N_{yx}}{N_{yx}+N_{xx}} \tag{30}$$

# 7. Simulation Results

We evaluated the performance of our proposed SIP-SegNet model both subjectively and objectively as explained in the following subsections.

## 7.1 Subjective Evaluation

Periocular components such as eyelashes, eyelids and spectacle reflection often obstruct the ocular region. Therefore, we analyzed the performance of our trained SIP-SegNet model subjectively against the various occlusion categories as shown in Table XII. We randomly picked the images from our testing dataset and arranged them into eight different categories consisting of 50 images each.

**Table XII: Categories of Ocular Occlusions for Subjective Evaluation.**

| Category | Description | Category | Description |
|---|---|---|---|
| **Clear** | This category refers to images without any obstruction in the ocular region. Sclera, iris and pupil regions in these images are clearly visible. 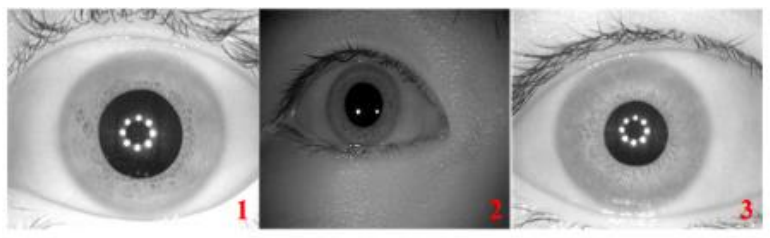 | **Eyelashes Occlusion** | This category of images refers to occlusion in the ocular region predominantly by the eyelashes as shown in the sample figures. The yellow bounding box represents the eyelash occlusion. 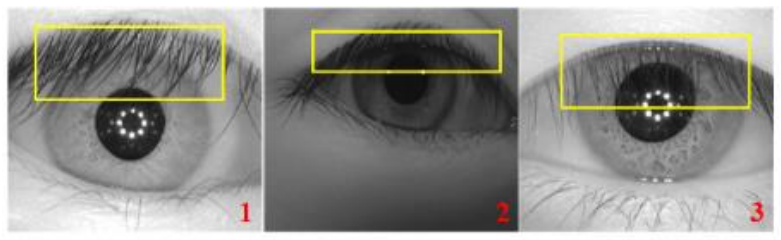 |
| **Upper Eyelid Occlusion** | The ocular regions in images of this category are partially occluded due to upper eyelid (blue bounding box). In sample images 1 and 2, the upper eyelid obstructs the iris and sclera region, while it can be observed from the sample figure 3 that upper eyelid occludes all the three regions (sclera, iris, pupil). 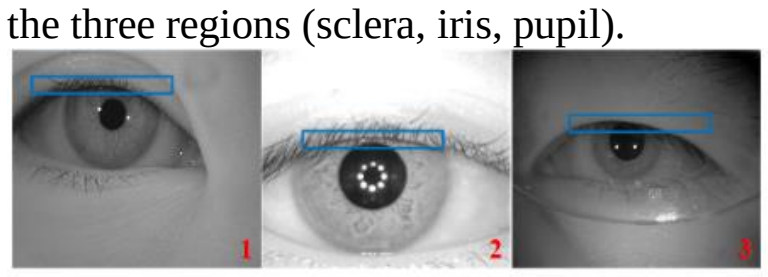 | **Spectacle Reflection** | This category refers to images with spectacle reflections obstructing the ocular region. These reflections can occur due to various reasons such as camera flash, light, etc. The spectacle reflections in the sample images are marked by asterisk (*). 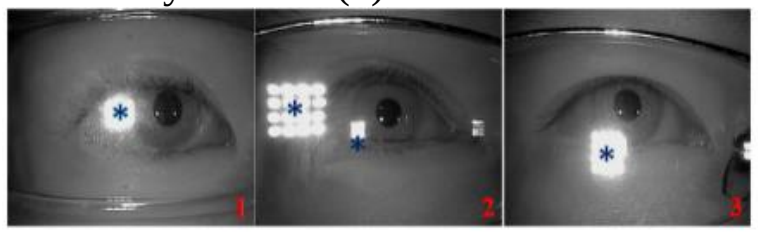 |
| **Lower Eyelid Occlusion** | The occlusion in these images is due to the lower eyelid as shown in the sample figures. The obstructed ocular region is represented with blue bounding box. 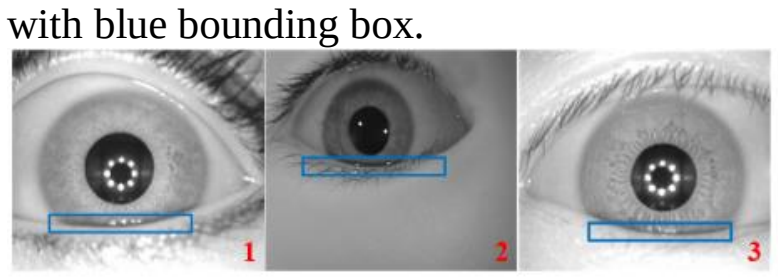 | **Poorly Acquired** | The ocular region in images of this category is not properly visible due to poor acquisition. 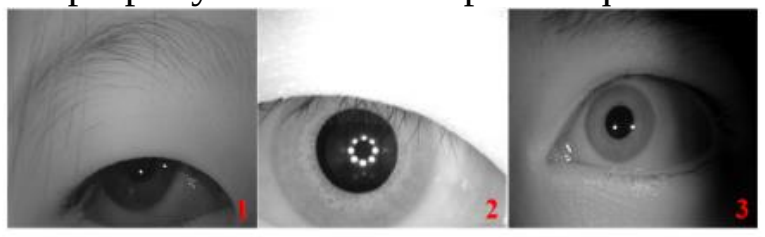 |
| **Upper Lower Eyelids Occlusion** | The occlusion in these images is caused by the upper and lower eyelids as represented with blue bounding box in the sample figures. 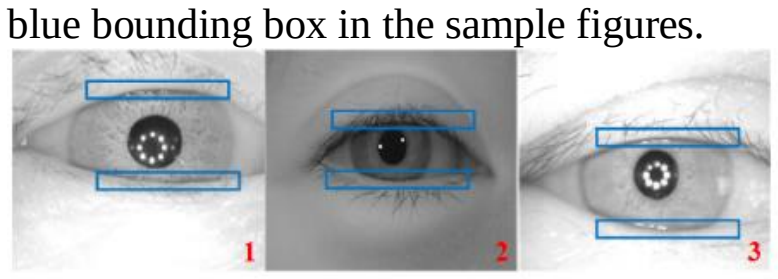 | **Heavily Occluded** | In the final category, we considered heavily occluded images, where the ocular region is obstructed due to multiple entities such as eyelashes, eyelids, reflections, spectacles etc. as shown in the sample images. 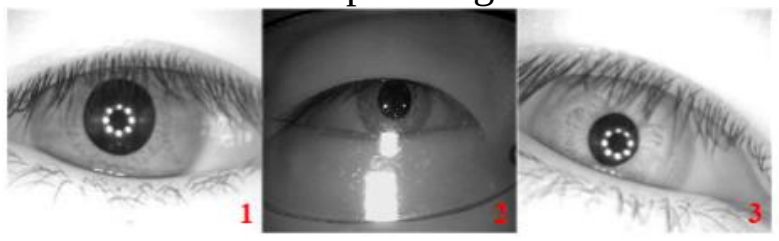 |

Fig. 21 shows the ocular segmentation results using SIP-SegNet model. Clear ocular images as shown in the first two rows of Fig. 21 does not contain any occlusion in the sclera, iris and pupil region. SIP-SegNet achieved the best results for images in this category, with segmented regions almost identical to the ground truth. The higher accuracy in such cases is perhaps due to discernible and clear contours of each ocular component.

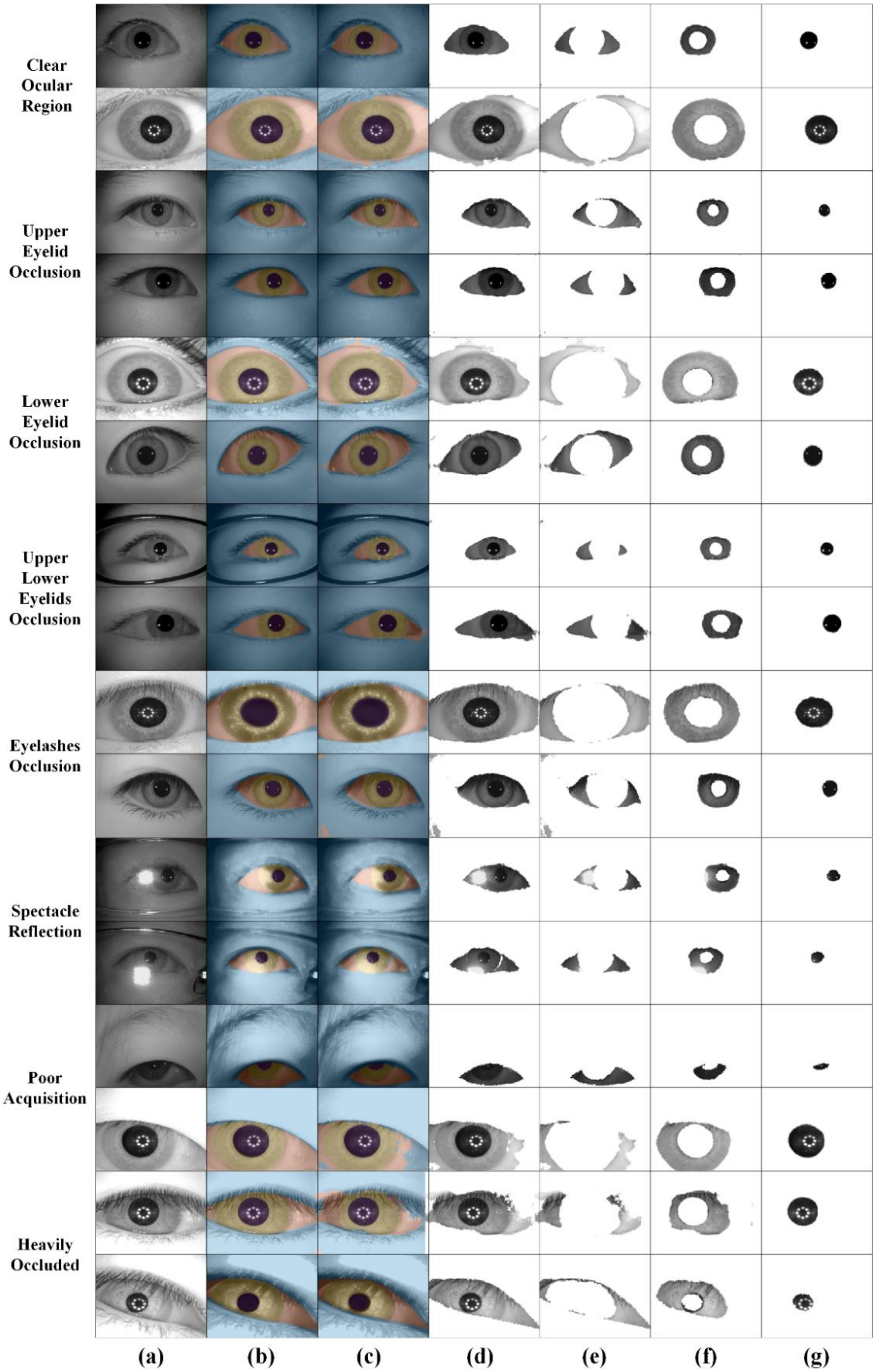

**Fig. 21: Evaluation of SIP-SegNet for ocular segmentation (a) Original image (b) Ground truth pixel labels overlaid on the original image (c) Segmented pixel labels using SIP-SegNet overlaid on the original image (d) Extraction of ocular region (e) Sclera region (f) Iris region and (g) Pupil region from the original image using segmented pixel labels.**

The next six rows present the segmentation results on images containing obstruction in the ocular region due to upper/ lower eyelid(s). This type of occlusion mostly disturbs the contour and symmetry of iris region. The proposed framework showed a good generalization on images in this category and classified the majority of ocular labels correctly with minor false positives. In addition, SIP-SegNet preserved the symmetry of ocular region as in the original image by precluding the obstructed sclera and iris regions as evident from the segmentation results. Similarly, our framework demonstrated good results for images predominated with eyelash occlusion and spectacle reflections. The proposed preprocessing technique for reflection removal and suppression of periocular region enabled SIP-SegNet to achieve higher segmentation accuracy in these images.

Further, we evaluated the performance of SIP-SegNet on poorly acquired images. We categorized images as poorly acquired if: (i) the ocular region is not captured completely due to poorly aligned shooting angle, (ii) ocular region is partially or completely dark due to shadow or illumination differences and (iii) ocular region contains motion blur due to movement of eye. The results validate that the proposed framework correctly segmented the iris and pupil region in most of the cases. However, SIP-SegNet showed deteriorated performance in segmenting sclera pixels of this category when compared with the other categories. The prime reason for this decline in performance is incomplete and irreconcilable contour of sclera in these images.

Finally, we computed the segmentation accuracy for images in heavily occluded category, where the ocular region is occluded through multiple obstructions such as eyelashes, eyelids, reflections etc. as shown in the last two rows of Fig. 21. The segmentation performance of SIP-SegNet on this set is not as accurate as it is for the other categories. Few false negatives against images in this category are also visible in the Fig. 21.

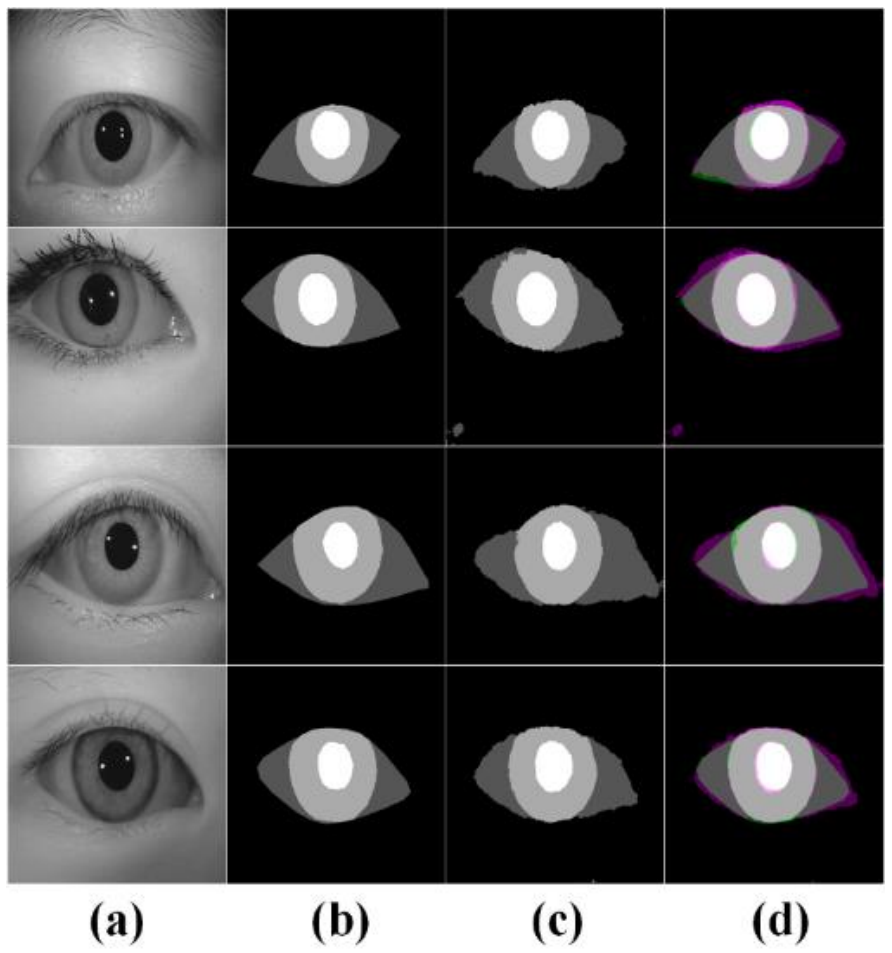

**Fig. 22: Ground truth versus predicted labels (a) Original image (b) Ground truth labels (c) Predicted labels (d) Overlapped ground truth and predicted labels. Magenta and green colors indicate the false positive and false negative ocular (sclera, iris, pupil) pixels respectively.**

The more intuitive representation of ocular segmentation results is shown in Fig. 22, where it can be observed that the SIP-SegNet predicted labels are almost identical to the ground truth. Visually, the segmented results of periocular, iris and pupil classes overlap well, however, the semantic segmentation results of sclera class are slightly less accurate. We measured the exact amount of overlap per class using the Jaccard index for each of the above discussed occlusion categories in the next subsection. Subjectively, the SIP-SegNet framework achieved the excellent segmentation results on both clear and occluded ocular images.

## 7.2 Objective Evaluation

The proposed SIP-SegNet framework is the first of its kind to perform joint segmentation of sclera, iris and pupil. We evaluated the performance of SIP-SegNet objectively via three main experiments as explained in the following subsections.

### 7.2.1 Evaluation on Test Datasets

SIP-SegNet is trained using augmented datastore containing a total of 31,220 training images from five different datasets of CASIA. In this first experiment, we compared the performance of SIP-SegNet on a total of 10,407 test images from all of the five CASIA datasets as shown in Table IX. We divided each of the five datasets into training, validation and testing sets as evident from the Table IX. The primary reason for this splitting is to analyze the network's generalization ability over various datasets in unconstrained environments. Secondly, adopting this splitting strategy allowed us to create five diverse test sets each with a varied number of test images. Lastly, this enabled us to evaluate new and distinct images, which the network never saw before i.e. during the training and validation phase.

The normalized confusion matrix generated on the complete test data containing 10,407 images is shown in Fig. 23. The diagonal row in the Fig. 23 represents the percentage of accuracy for correctly classified pixels in each class.

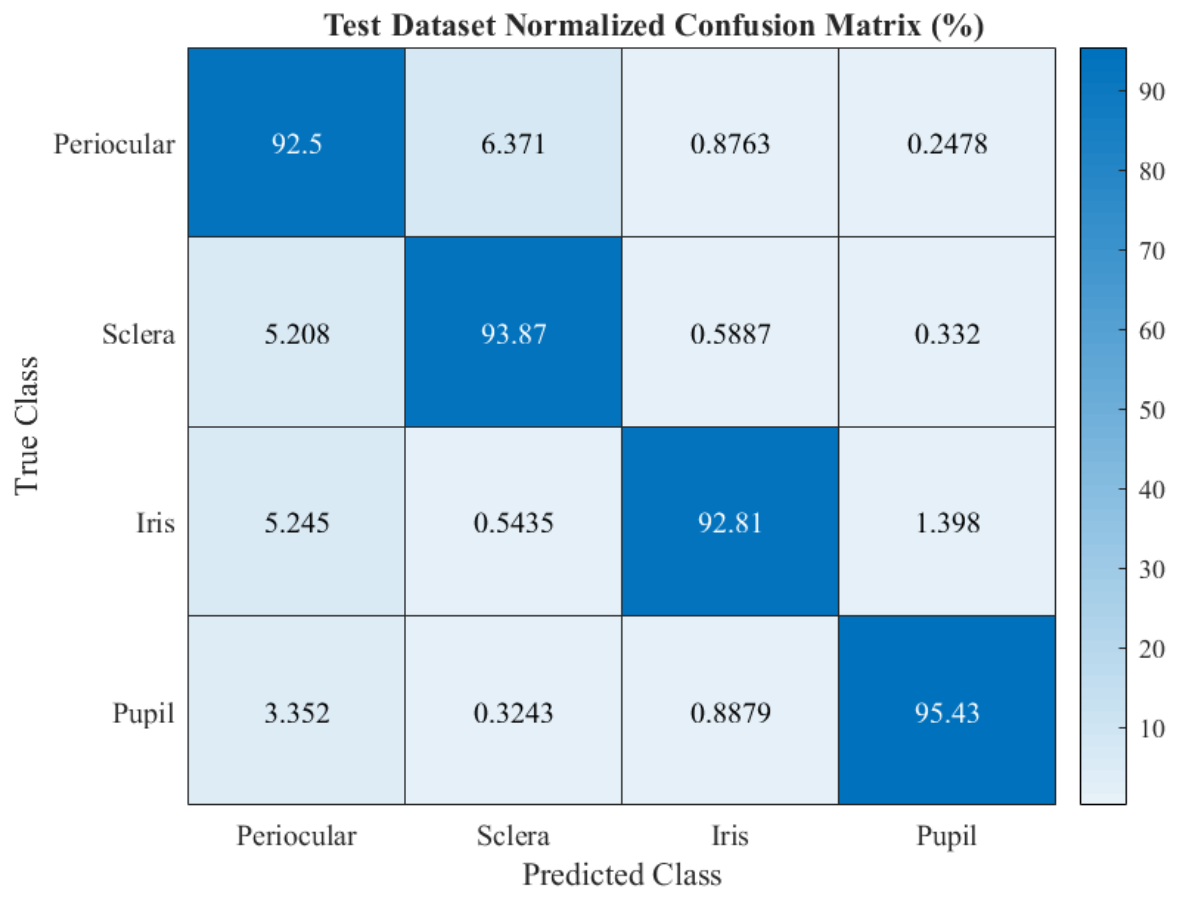

**Fig. 23: Confusion matrix of complete test data.**

The numerical results in Table XIII show the performance of SIP-SegNet against individual testing dataset. Each value in the Table XIII represents the average value computed over the total number of test images in each dataset. In addition, we presented the class wise results, where the value in bold indicates the optimal value for each evaluation metric with respect to class and dataset. From the Table XIII, it can be observed that our proposed framework achieved higher accuracy in terms of correctly segmenting labels for each class. Also, we can analyze that the SIP-SegNet performed better on the CIS and CIL datasets as compared to other three datasets (CII, CITW, CIT). There can be multiple reasons for this decline in the SIP-SegNet performance in these datasets. CII dataset contains the close up eye images, while all the other datasets contain images captured from a distance. Also, the ratio of CII images is low as compared to other datasets. Conversely, CIT contains the most challenging images with a lot more intra-class variations such as eyeglasses and specular reflections. Moreover, the CIT dataset with 20000 images is the largest one and obtaining the highest accuracy for such a large dataset is a challenging task. Further, the images in CITW dataset were acquired during the annual festival in Beijing and it is the only dataset where images are captured in outdoor environment. The illumination difference of this dataset made it totally unique as compared to other datasets and this is the reason our proposed model achieved the least accuracy for CITW. However, it is pertinent to mention here that despite these variations in the datasets, there is only a minor difference in the accuracy value of each dataset, which affirms the robustness and reliability of SIP-SegNet framework irrespective of the datasets.

In addition, Table XIII shows the comparatively low values of precision and recall especially for periocular class, which indicates the uncertainty of the model against predicting the positive periocular labels. Similarly, the higher NPV and specificity values for each label in all datasets affirms that the SIP-SegNet is able to rule out the negative instances correctly. The higher values of dice coefficient and IoU for both pupil and iris describes the higher overlapped region and similarity index between the ground truth and predicted labels of these classes. Whereas, the same metrics gave relatively low values for periocular and sclera classes. This can be due to the fact that the CASIA datasets are in grayscale, where the pixel intensities in the periocular and sclera region are relatively close to each other and are not distinct as the other two labels (iris and pupil). As a result, the network performance slightly declined in predicting the periocular and sclera labels. Similarly, the higher values of nice2 for both periocular and sclera class reflect the higher ratio of false predicted labels in these classes when compared to iris and pupil.

**Table XIII: SIP-SegNet Evaluation using Various Metrics.**

| Evaluation Metrics | Classes | Datasets | | | | |
|---|---|---|---|---|---|---|
| | | **CII** | **CIS** | **CIL** | **CITW** | **CIT** |
| **Accuracy (%)** | Periocular | 94.42 | 94.88 | **94.93** | 94.66 | 94.49 |
| | Sclera | 96.51 | **96.91** | 96.87 | 96.57 | 96.43 |
| | Iris | 97.71 | **97.79** | 97.78 | 97.44 | 97.35 |
| | Pupil | 98.44 | 98.55 | **98.63** | 98.19 | 98.00 |
| | **All Classes** | 96.77 | 97.03 | **97.05** | 96.71 | 96.57 |
| **Precision (%)** | Periocular | 86.61 | **87.55** | 87.51 | 86.75 | 86.65 |
| | Sclera | 92.45 | 93.15 | **93.35** | 92.80 | 92.46 |
| | Iris | 97.74 | 97.93 | **97.95** | 97.28 | 96.73 |
| | Pupil | 98.13 | **98.33** | 98.32 | 97.65 | 97.41 |
| | **All Classes** | 93.73 | 94.24 | **94.28** | 93.62 | 93.31 |
| **Recall (%)** | Periocular | 91.89 | 92.71 | **92.98** | 92.81 | 92.14 |
| | Sclera | 93.70 | **94.58** | 94.21 | 93.52 | 93.34 |
| | Iris | 93.01 | **93.14** | 93.05 | 92.32 | 92.54 |
| | Pupil | 95.58 | 95.84 | **96.18** | 95.05 | 94.53 |
| | **All Classes** | 93.54 | 94.07 | **94.10** | 93.43 | 93.14 |
| **Specificity (%)** | Periocular | 95.27 | **95.61** | 95.57 | 95.28 | 95.27 |
| | Sclera | 97.45 | 97.68 | **97.76** | 97.58 | 97.46 |
| | Iris | 99.28 | 99.34 | **99.35** | 99.14 | 98.96 |
| | Pupil | 99.39 | **99.46** | 99.45 | 99.24 | 99.16 |
| | **All Classes** | 97.85 | 98.02 | **98.03** | 97.81 | 97.71 |
| **NPV (%)** | Periocular | 94.24 | 97.52 | **97.61** | 97.54 | 97.32 |
| | Sclera | 97.89 | **93.18** | 98.06 | 97.84 | 93.77 |
| | Iris | 97.71 | **97.75** | 96.72 | 97.48 | 97.55 |
| | Pupil | 98.54 | 98.62 | **98.74** | 97.37 | 98.19 |
| | **All Classes** | 97.10 | 96.77 | **97.78** | 97.56 | 96.71 |
| **IoU (%)** | Periocular | 80.46 | 81.91 | **82.08** | 81.29 | 80.69 |
| | Sclera | 87.04 | **88.43** | 88.28 | 87.19 | 86.74 |
| | Iris | 91.05 | **91.35** | 91.27 | 90.00 | 89.74 |
| | Pupil | 93.87 | 94.30 | **94.62** | 92.93 | 92.21 |
| | **All Classes** | 88.11 | 89.00 | **89.07** | 87.85 | 87.34 |
| **Dice (%)** | Periocular | 89.17 | 90.05 | **90.16** | 89.68 | 89.31 |
| | Sclera | 93.07 | **93.86** | 93.78 | 93.16 | 92.90 |
| | Iris | 95.32 | **95.48** | 95.44 | 94.74 | 94.59 |
| | Pupil | 96.84 | 97.07 | **97.24** | 96.34 | 95.95 |
| | **All Classes** | 93.60 | 94.11 | **94.15** | 93.48 | 93.19 |
| **Nice2 (%)** | Periocular | 6.42 | 5.84 | **5.72** | 5.96 | 6.30 |
| | Sclera | 4.42 | **3.87** | 4.01 | 4.45 | 4.60 |
| | Iris | 3.85 | **3.76** | 3.80 | 4.27 | 4.25 |
| | Pupil | 2.51 | 2.35 | **2.18** | 2.85 | 3.16 |
| | **All Classes** | 4.30 | 3.96 | **3.93** | 4.38 | 4.57 |

NPV= Negative Predictive Value, IoU= Intersection over Union. The value in bold indicates the optimal value for each evaluation metric with respect to class and dataset.

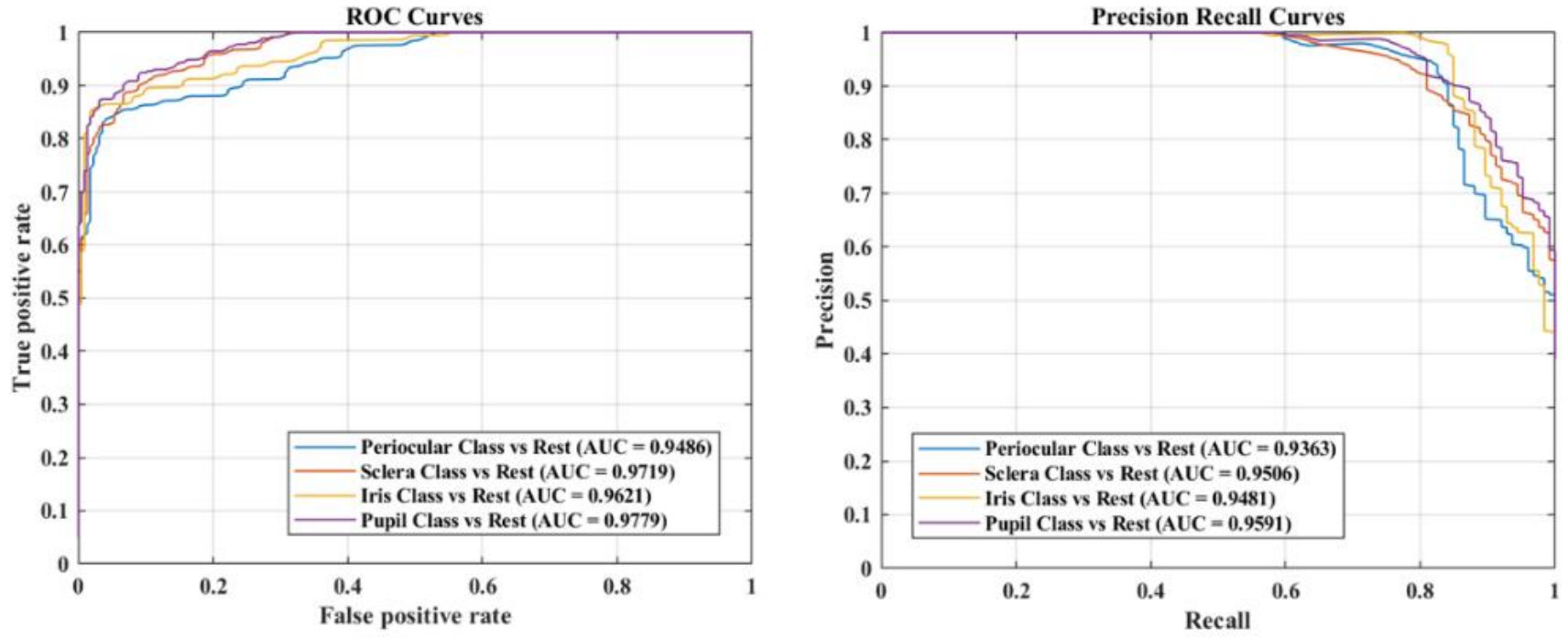

(a) (b)

**Fig. 24: (a) Receiver operating characteristic (ROC) curve for each class (b) Precision-recall curve for each class.**

Further, we analyzed the performance of SIP-SegNet using the receiver operating characteristic (ROC) and precision-recall curves of each class as shown in Fig. 24. We generated the ROC and precision-recall curves for each class by varying the pixel classification threshold between 0 and 1 with a step-size of 0.003. It can be observed from the ROC curves in Fig. 24(a) that our proposed SIP-SegNet model is trained skillfully in correctly predicting the positive labels in each class. The higher area under the curve (AUC) value for each class also quantifies the superior predictive performance of SIP-SegNet model. Since, in our study, there is a huge imbalance between the classes, therefore, precision-recall curves show the advantages of SIP-SegNet algorithm more intuitively due to absence of true negatives in precision and recall equations. The higher AUC values for each class shows that the SIP-SegNet is trained precisely and returns accurate results (high precision) and predicted most of the positive results correctly (high recall).

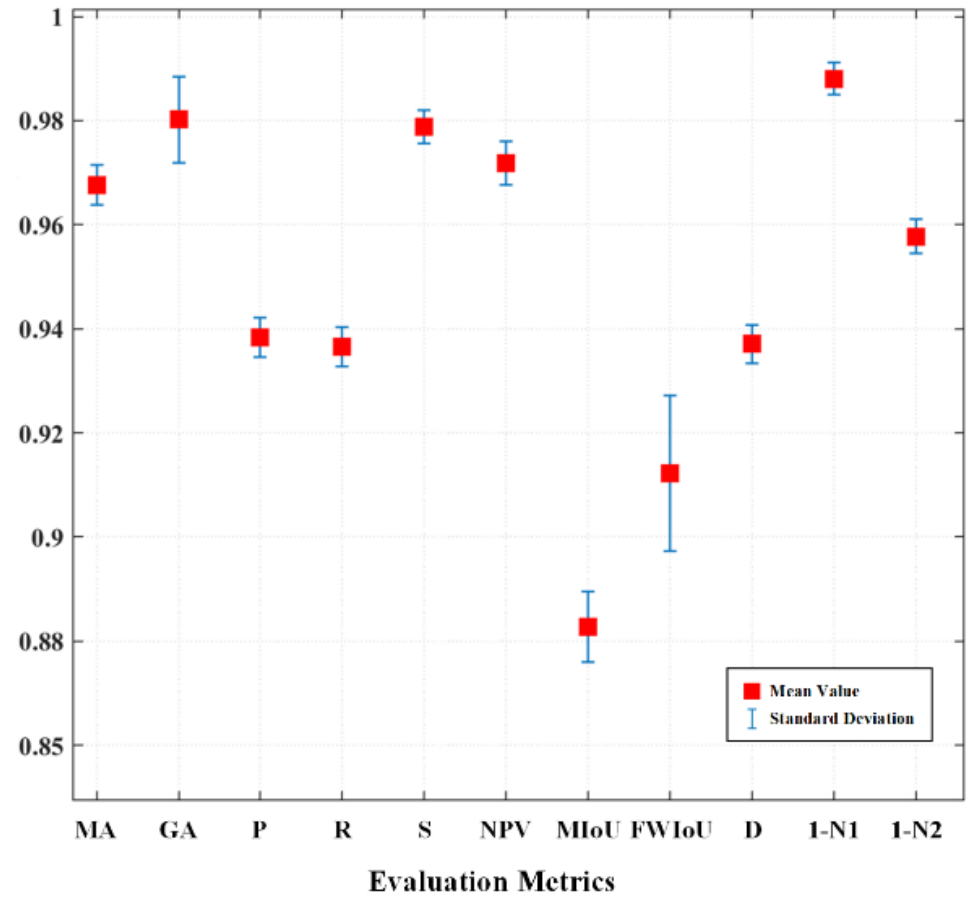

**Fig. 25: SIP-SegNet performance evaluation over all classes all datasets. MA= Mean Accuracy, GA= Global Accuracy, P= Precision, R= Recall, S= Specificity, NPV= Negative Predicted Value, MIoU= Mean Intersection over Union, FWIoU= Frequency Weighted Intersection over Union, D= Dice Similarity Coefficient, N1= Nice1 and N2= Nice 2.**

The evaluation of SIP-SegNet model for all test images regardless of the class is shown in Fig. 25. The red square in the Fig. 25 represents the mean value of each evaluation metric, which is computed over the total number of test images and considering all classes (periocular, sclera, iris and pupil). The length of the blue line above and below the red square indicates the standard deviation of each evaluation metric. It can be observed that the mean value of each metric is found to be greater than 90% except for MIoU, which has a mean value of 88.27%. SIP-SegNet achieved the mean value of specificity and NPV greater than 97% for all test images, whereas the mean value of precision and recall are less than 94%. This shows that our proposed framework predicted the negative labels comparatively better than the positive ones. Similarly, the mean value of 1.8% for nice1 indicates the higher similarity between the ground truth and predicted labels.

### 7.2.2 Evaluation based on Ocular Occlusion Categories

In this experiment, we analyzed the performance of our proposed SIP-SegNet framework objectively based on different ocular occlusion categories as discussed earlier. We used 50 randomly selected images against each category to evaluate the performance of SIP-SegNet as shown in the Fig. 26.

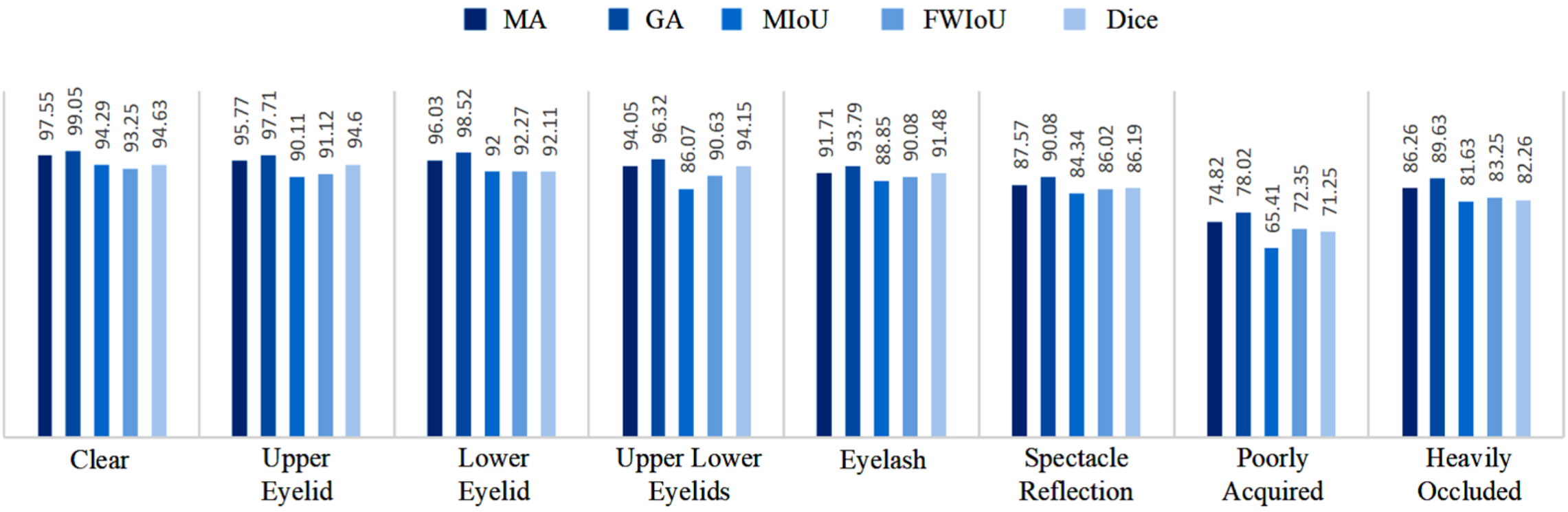

**Fig. 26: SIP-SegNet performance evaluation for all classes against various occlusion categories. MA= Mean Accuracy, GA= Global Accuracy, MIoU= Mean Intersection over Union, FWIoU= Frequency Weighted Intersection over Union, D= Dice Similarity Coefficient.**

As evident from the Fig. 21 and Fig. 26, the SIP-SegNet performed best on the clear images, followed by the eyelid and eyelash occlusion category images. Whereas, its performance relatively declined for the category of spectacle reflection and heavily occluded images. Further, the proposed framework performed least accurate on poorly acquired images. The last three categories can be termed as the challenging cases in our study. Also, the number of images belonging to these categories are fewer in the total dataset. Therefore, this scarcity of data led to an improper training of network to handle such exceptions.

#### 7.2.3 Comparison with state-of-the-art Literature

In this experiment, we compared the performance of our proposed SIP-SegNet model with the existing state-of-the-art segmentation frameworks. To the best of our knowledge, there is no reported work in literature which focuses on the joint semantic segmentation of sclera, iris and pupil. Therefore, we first compared our sclera segmentation results with the existing algorithms that work towards the segmentation of sclera, followed by pupil and iris.

##### 7.2.3.1 Evaluation of Sclera Segmentation

The number of deep learning algorithms for segmentation of sclera are few. Moreover, none of these algorithms utilized the CASIA database, therefore, the direct comparison between our sclera segmentation results and existing frameworks is not possible. However, we have compared our sclera segmentation results against various evaluation metrics over different algorithms and databases as shown in Table XIV.

**Table XIV: Comparison of SIP-SegNet Sclera Segmentation with State-of-the-art Algorithms. Results are represented as Mean ± Standard Deviation.**

| Algorithm | Database | Test Images | Accuracy (%) | Specificity (%) | Precision (%) | Recall (%) | F1-Score (%) | Dice (%) | IoU (%) | Nice1 (%) | Nice2 (%) |
|---|---|---|---|---|---|---|---|---|---|---|---|
| UNet [62] | SBVPI | 698 | - | - | 93.60 ± 4.40 | 93.00 ± 3.70 | 93.30 ± 3.70 | - | - | - | - |
| RefineNet-50 [62] | | | - | - | 95.90 ± 2.00 | 95.90 ± 2.00 | 95.90 ± 1.80 | - | - | - | - |
| RefineNet-101 [62] | | | - | - | 95.30 ± 2.50 | 95.10 ± 2.30 | 95.20 ± 2.10 | - | - | - | - |
| SegNet [62] | | | - | - | 94.90 ± 2.40 | 94.90 ± 2.20 | 94.90 ± 2.10 | - | - | - | - |
| FCN* [63] | UBIRIS v2 | 200 | - | - | 88.45 ± 6.98 | 87.31 ± 6.68 | 87.48 ± 3.90 | - | - | - | - |
| FCN* [63] | MICHE-I | 400 | - | - | 89.90 ± 9.82 | 87.59 ± 11.28 | 88.32 ± 9.80 | - | - | - | - |
| CNN [23] | MASD | 30 | 81.10 | - | 51.30 | 48.70 | - | - | - | - | - |
| CNN+CRF [23] | MASD | 30 | 83.20 | - | 54.70 | 74.30 | - | - | - | - | - |
| Sclera-Net^ [64] | SBVPI | 2399 | - | - | 94.40 ± 3.28 | 98.17 ± 1.60 | 96.24 ± 1.71 | - | 91.80 | 0.93 | - |
| CNNSRE [21] | MASD | 734 | 87.65 | - | 86.00 | 86.00 | - | - | - | - | - |
| **SIP-SegNet** | **CASIA v4** | **10407** | **96.66 ± 0.19** | **97.59 ± 0.12** | **92.84 ± 0.36** | **93.87 ± 0.46** | **93.35 ± 0.39** | **93.35 ± 0.39** | **87.54 ± 0.69** | **1.27 ± 0.09** | **4.27 ± 0.28** |

*Authors in [63] used multiple algorithms, results here represent their best algorithm, ^authors in [64] performed two-fold cross validations, results here show the average values.

##### 7.2.3.2 Evaluation of Iris Segmentation

Furthermore, SIP-SegNet is evaluated based on segmentation of iris with other existing schemes in literature as shown in Table XV. It is pertinent to mention here that the comparison in Table XV is drawn over the same dataset i.e. CASIA iris interval. In this context, we assessed the performance of SIP-SegNet with various benchmark iris segmentation algorithms including GST [103], a generalized structure tensor based iris boundary segmentation framework; OSIRIS [104], a circular Hough transform based method with boundary refinement; WAHET [105], an adaptive multi-scale Hough transform segmentation algorithm based on elliptical models; IFPP [106], a framework based on iterative Fourier-series approximation with pulling and pushing techniques; CAHT [107], a circular Hough transform based traditional sequential method with contrast enhancement; MASEK [108], which is an iris pattern based biometric identification system and IDO [33], the Daugman's pioneering work towards iris recognition based on integro-differential operator.

**Table XV: Comparison of SIP-SegNet Iris Segmentation with previous methods using CASIA Iris Interval Dataset. Results are represented as Mean ± Standard Deviation.**

| Algorithm | Accuracy (%) | Specificity (%) | Precision (%) | Recall (%) | F1-Score (%) | Dice (%) | IoU (%) | Nice1 (%) | Nice2 (%) |
|---|---|---|---|---|---|---|---|---|---|
| GST [103] | - | - | 89.91 ± 7.37 | 85.19 ± 18 | 86.16 ± 11.53 | - | - | - | - |
| Osiris [104] | - | - | 93.03 ± 4.95 | 97.32 ± 7.93 | 89.85 ± 5.47 | - | - | 05.35 ± 02.40^ | - |
| WAHET [105] | - | - | 85.44 ± 9.67 | 94.72 ± 9.01 | 89.13 ± 8.39 | - | - | 6.08† | 8.42† |
| IFFP [106] | - | - | 83.50 ± 14.26 | 91.74 ± 14.74 | 86.86 ± 13.27 | - | - | - | - |
| CAHT [107] | - | - | 82.89 ± 9.95 | 97.68 ± 4.56 | 89.27 ± 6.67 | - | - | 11.61† | 14.70† |
| MASEK [108] | - | - | 89.00 ± 6.31 | 88.46 ± 11.52 | 88.30 ± 7.99 | - | - | - | - |
| IDO [33] | - | - | 61.62 ± 18.71 | 71.34 ± 22.86 | 65.61 ± 19.96 | - | - | 1.15ξ | - |
| IrisSeg [43] | - | - | 92.15 ± 3.34 | 94.26 ± 4.18 | 93.10 ± 2.65 | - | - | 02.85 ± 01.62^ | - |
| IrisDenseNet [44] | - | - | 98.10 ± 1.07 | 97.10 ± 2.12 | 97.58 ± 0.99 | - | - | - | - |
| FCN [45] | - | - | - | - | 97.90 ± 00.68 | - | - | 01.15 ± 00.37 | - |
| GAN [45] | - | - | - | - | 96.13 ± 05.35 | - | - | 01.45 ± 03.71 | - |
| PixISegNet [49] | 97.86 | - | 97.89 | 97.90 | 97.90 | - | 95.87 | 0.01 | - |
| FCN [46] | - | - | - | - | 98.60 | - | - | 0.79 | 0.88 |
| FCEDNs-1 [47] | - | - | - | - | 88.26 | - | - | 5.61 | 5.88 |
| FCEDNs-2 [47] | - | - | - | - | 90.72 | - | - | 4.48 | 4.38 |
| FCEDNs-3 [47] | - | - | - | - | 91.92 | - | - | 3.91 | 4.07 |
| U-Net [48] | - | - | - | - | 97.23 | - | - | 0.61 | 1.27 |
| Dense U-Net [48] | - | - | - | - | 97.07 | - | - | 0.31 | 1.22 |
| **SIP-SegNet** | **97.61 ± 0.18** | **99.21 ± 0.15** | **97.53 ± 0.47** | **92.81 ± 0.32** | **95.11 ± 0.37** | **95.11 ± 0.37** | **90.68 ± 0.68** | **1.41 ± 0.08** | **3.99 ± 0.23** |

The results of GST [103], Osiris [104], WAHET [105], IFFP [106], CAHT [107], MASEK [108] and IDO [33] are referred from [43],[44]; ^the values of Nice1 in Osiris [104] and IrisSeg [43] are referred from [45]; †the values of Nice1 and Nice2 in WAHET [105] and CAHT [107] are referred from [46]; ξthe value of Nice1 in IDO [33] is referred from [49].

### 7.2.3.3 Evaluation of Pupil Segmentation

Similarly, we compared the performance of our SIP-SegNet in terms of pupil segmentation with several existing algorithms as shown in Table XVI.

**Table XVI: Comparison of SIP-SegNet Pupil Segmentation with State-of-the-art Algorithms. Results are represented as Mean ± Standard Deviation.**

| Algorithm | Database | Test Images | Accuracy (%) | Specificity (%) | Precision (%) | Recall (%) | F1-Score (%) | Dice (%) | IoU (%) | Nice1 (%) | Nice2 (%) |
|---|---|---|---|---|---|---|---|---|---|---|---|
| DeepVOG^ [66] | IITD | 763 | - | - | - | - | - | 97.80 | - | - | - |
| IPSegNet1* [75] | CASIA | 7000 | 99.48 | - | 99.48 | 98.22 | - | - | - | - | - |
| IPSegNet2* [75] | CASIA | 7000 | 99.84 | - | 99.84 | 99.96 | - | - | - | - | - |
| BSOM^ [73] | CASIA | 2655 | 97.95 | - | - | - | - | - | - | - | - |
| U-Net^ [71] | UBIRIS v2 | 2250 | - | - | 77.30 | 90.20 | 81.50 | - | - | - | - |
| BSOM+CA ^ [74] | CASIA | 10,000 | 99.85 | - | - | - | - | - | - | - | - |
| DeepEye [76] | Multiple | - | 92.10 | 99.90 | - | 89.60 | - | - | - | - | - |
| **SIP-SegNet** | **CASIA v4** | **10407** | **98.36 ± 0.23** | **99.34 ± 0.12** | **97.97 ± 0.37** | **95.44 ± 0.58** | **96.69 ± 0.48** | **96.69 ± 0.48** | **93.59 ± 0.89** | **1.03 ± 0.07** | **2.61 ± 0.35** |

^tested on multiple datasets, the values here represent the best results achieved, *the results in [75] are presented for three Intersection over Union (IoU) thresholds (0.35,0.50,0.65). The values in this table refer to IoU threshold = 0.50.

It can be observed from the results in Table XIV to Table XVI that the performance of the proposed SIP-SegNet framework is consistent with the state-of-the-art algorithms. It is pertinent to mention that the SIP-SegNet is a novel algorithm which performs the joint semantic segmentation of periocular region, sclera, iris and pupil simultaneously unlike all the existing algorithms, which deal with one or two ocular modalities only. Also, the performance of SIP-SegNet model is evaluated on a large number of subjects (test images) when compared with existing algorithms. These images are from five different CASIA datasets (CII, CIS, CIL, CITW, CIT) containing a lot more intra-class variations (spectacle reflections, illumination differences, occlusions, distance etc.). Furthermore, these different datasets and intra-class variations created a huge imbalance between the four classes (periocular, sclera, iris and pupil) and particularly more biasness towards the periocular class as shown in Fig. 18. However, despite these foregoing factors, SIP-SegNet performed extremely well and the segmentation results of each class are comparable and consistent with the existing frameworks. In addition, our simulation results validate the robustness and reliability of SIP-SegNet framework irrespective of the datasets and variations.

## 8. Discussion

The motivation of this work is to segment multiple ocular traits simultaneously towards devising a multimodal ocular segmentation framework. We used Segnet [32] in this work, which is a type of CNN designed for semantic image segmentation. There are multiple reasons for using Segnet such as: (i) it is a lot easier to train end to end, (ii) Segnet obviates the need of high computational resources due to its small number of parameters unlike in DeconvNet [109] and (iii) it reuses the pooling indices only in decoder path and requires far less memory as compared to U-Net network [110], which consumes high memory due to transfer of complete feature maps from the encoders to the corresponding decoders. Moreover, in our SIP-SegNet model, we changed the fully connected layers in the encoder path with the convolutional layers. This resulted in two major improvements in the model: (i) the higher resolution feature maps are preserved at the deepest encoder outputs and (ii) the number of parameters in the encoder path of our model reduced significantly from 134 million to 14.7 million.

The periocular components as evident from the Fig. 1 can greatly affect the performance of the ocular segmentation framework. Moreover, for several images in the datasets we observed that: (i) the difference between the black intensities in the ocular region is very small as evident from Fig. 8 and (ii) the intensities are clustered predominantly around the lower or middle range as evident from Fig. 9. Therefore, in this work we enhanced the ocular traits using preprocessing stage and subsequently suppressed the information of the periocular components. Additionally, we removed the various reflections in the pristine image via improved holes filling strategy as shown in Fig. 11. These preprocessing techniques enabled SIP-SegNet to train more precisely and converge quickly.

Further, the classes in the training data are highly imbalanced, with periocular being the dominant class having around 75% of the labels. On the contrary, pupil class had the least i.e. less than 5% of the labels as shown in Fig. 8. This kind of biasness can lead to false segmentation of marginalized classes despite achieving greater training accuracy, as the network only learnt to classify the dominant class accurately. Therefore, in our proposed framework, we used the inverse frequency weighting approach to balance the classes by assigning increased weights to the underrepresented classes (sclera, iris and pupil). The higher AUC values for each class as shown in Fig. 24(b) validates the superior performance of SIP-SegNet in terms of the accurate results (high precision) and the correct prediction of most of the positive labels (high recall).

In addition, we applied various augmentations on the training data as shown in Fig. 15. This enabled the network to train and handle images with various perspective more precisely. We further ensured the proper and robust training of SIP-SegNet model by specifying a high learning rate. To prevent the gradients of the network to grow uncontrollably, we enabled gradient clipping and set the gradient threshold equal to 0.005.

We have conducted a number of experiments to evaluate the performance of SIP-SegNet framework. We computed the segmentation accuracy of each class for various occlusion categories as shown in Fig. 21 and Fig. 26. Moreover, the proposed framework is tested over multiple challenging datasets using various evaluation metrics as shown in Table XIII and Fig. 25. Furthermore, the comparative analysis between the proposed SIP-SegNet and existing state-of-the-art algorithms is presented in Table XIV to Table XVI. The demonstrated simulation results in this paper authenticate the optimal performance of our proposed framework. In addition, the confusion matrix as shown in Fig. 23 validates that the network trained optimally and achieved accuracy greater than 90% for each class.

The SIP-SegNet training stage is time consuming, which can be referred as a limitation. This is due to dense network connectivity. Further, the size of the mini-batch can be another constraint, which should be kept low to avoid the high memory consumption. However, the testing phase of SIP-SegNet is quite fast once the network is fully trained. Although our proposed model achieved comparable results with the mean F1 score of 89.67, 93.35, 95.11 and 96.69 for the periocular, sclera, iris and pupil classes respectively, the segmentation accuracy of SIP-SegNet can further be improved by incorporating various post-processing techniques.

In the future, the proposed model can be further refined using additional publicly available datasets with more challenging and unique ocular images. In addition, we can explore the potential of different deep learning architectures to further increase the system performance over time.

## 9. Conclusion

In this paper, we presented a novel joint semantic segmentation framework called SIP-SegNet for segmentation of ocular traits including sclera, iris and pupil. SIP-SegNet is a fully convolutional encoder-decoder network, where the encoder path consisting of 13 convolutional layers and 5 max pooling operations, reduces the spatial resolution to 1/32 times the pristine image. Whereas, the decoder path upsamples the reduced image in a step-wise manner using the pooling indices of the corresponding encoders. Multi-class soft-max classifier is used at the end of the decoder path for pixel-wise classification. SIP-SegNet is first of its kind to segment all three ocular traits (sclera, iris and pupil) simultaneously. Further, the proposed framework incorporates a useful preprocessing stage to effectively denoise and enhance the pristine images. Moreover, the periocular information in the enhanced image is also suppressed using fuzzy filtering technique, since the appearance of various occlusions such as eyelids, eyelashes, specular reflections etc. can greatly affect the ocular segmentation accuracy. In our work, we applied the empirical training methodology with varying training hyper parameters to increase the generalization ability of SIP-SegNet model and to restrict the overfitting problem during the training stage. We tested the effectiveness and robustness of SIP-SegNet on five challenging datasets. The proposed framework accurately segments the ocular modalities irrespective of intra-class variations in these datasets. The experimental results demonstrate the superiority of SIP-SegNet with the mean f1 score of 93.35, 95.11 and 96.69 for the sclera, iris and pupil classes respectively. In future, SIP-SegNet framework can be extended towards developing a robust multimodal biometric recognition system based on ocular traits.

## Declaration of Competing Interest

The authors declare that they have no known competing financial interests or personal relationships that could have appeared to influence the work reported in this paper.

## CRediT Authorship Contribution Statement

**Bilal Hassan and Ramsha Ahmed:** Conceptualization, Methodology, Software, Investigation, Writing - Original Draft, Writing - Review & Editing and Visualization. **Taimur Hassan:** Conceptualization, Software, Investigation, Writing - Original Draft, Writing - Review & Editing and Visualization. **Naoufel Werghi:** Investigation, Validation, Formal analysis, Writing – Review, Resources and Supervision.

## Acknowledgement

We are grateful to Chinese Academy of Sciences Institute of Automation for making their datasets publicly available for research purposes. We have used five subsets of the CASIA-IrisV4 database to perform this study [80].